\documentclass{article}
\usepackage{iclr2027_conference,times}

\usepackage[utf8]{inputenc}
\usepackage[T1]{fontenc}

\usepackage{amsmath,amsfonts,bm}

\def\eqref#1{equation~\ref{#1}}

\def\1{\bm{1}}

\DeclareMathAlphabet{\mathsfit}{\encodingdefault}{\sfdefault}{m}{sl}
\SetMathAlphabet{\mathsfit}{bold}{\encodingdefault}{\sfdefault}{bx}{n}

\usepackage{hyperref}
\usepackage{url}
\usepackage{booktabs}
\usepackage{amsmath}
\usepackage{amsfonts}
\usepackage{algorithm}
\usepackage{algpseudocode}
\usepackage{float}
\usepackage{graphicx}
\usepackage{subcaption}
\usepackage{multirow}
\usepackage{makecell}      %
\usepackage{needspace}     %
\usepackage{wrapfig}       %
\usepackage{enumitem}      %
\usepackage{placeins} %
\usepackage[breakable,skins]{tcolorbox} %
\newtcolorbox{examplebox}[2][]{enhanced, breakable, title={#2},
  colback=white, colframe=black!30, colbacktitle=black!65, coltitle=white,
  fonttitle=\small\bfseries\itshape,
  fontupper=\small\ttfamily\raggedright\frenchspacing, fontlower=\small,
  boxrule=0.5pt, arc=1.5mm, left=5pt, right=5pt, top=4pt, bottom=4pt,
  segmentation style={black!30, dashed}, #1}
\usepackage{microtype}
\usepackage{xcolor}
\usepackage{caption}
\graphicspath{{figures/}}
\usepackage{titlesec}

\renewcommand{\arraystretch}{0.6}

\titlespacing\section{0pt}{1.8ex plus 0.0ex minus 0.3ex}{1.3ex plus 0.0ex minus 0.2ex}
\titlespacing\subsection{0pt}{1.8ex plus 0.1ex minus 0.3ex}{1.2ex plus 0.1ex}
\titlespacing\paragraph{0pt}{-0.3ex}{1.1ex plus 0.3ex}


\title{Curriculum Learning Teaches LLMs Long-Horizon Implicit Reasoning, Up To A Limit}

\author{%
  Jimson Huang\thanks{Equal contribution.} \quad Maxwell Nickell\footnotemark[1] \quad Ruowang Zhang \quad Abulhair Saparov \\
  Purdue University\\
  West Lafayette, IN 47907, USA \\
  \texttt{\{huan2073, mnickel, zhan5763, asaparov\}@purdue.edu} \\
}

\iclrfinalcopy

\begin{document}
\maketitle

\begin{abstract}
LLMs ``reason'' by externalizing intermediate steps as tokens through chain-of-thought (CoT), yet even frontier reasoning models hit abrupt performance cliffs on long-horizon tasks. \emph{Implicit reasoning}---reasoning within a single forward pass---is useful during CoT generation, as the model can ``search ahead'' to decide the correct next move without explicit search. Models with weak implicit reasoning are more likely to commit to wrong paths and must attempt to recover via backtracking, while models with strong implicit reasoning identify the correct proof strategy directly, shortening the CoT. However, prior work shows LLMs struggle to reason implicitly beyond shallow depths, with no relief from scale. We investigate whether curriculum learning can extend this depth by fine-tuning on examples with progressively increasing vocabulary size and \emph{lookahead}---the number of steps a model must search ahead to identify the correct next move. At matched compute, curriculum-trained models achieve lookaheads where no-curriculum training fails---\texttt{Qwen3-0.6B} reaches $98\%$ accuracy on problems with lookahead up to $256$ with curriculum, compared to $45\%$ without. However, progress slows as fine-tuning continues, following a saturating exponential that suggests a \emph{maximum intrinsic lookahead} to which models can learn to reason implicitly. Larger models reach higher fitted ceilings, fine-tuning pretrained models far outpaces training from scratch, and curriculum step sizes that are too large stall progression. Finally, the capability acquired on synthetic data transfers to natural language reasoning, with curriculum-trained models outperforming models trained without curriculum at matched compute on $11$ natural language reasoning benchmarks. 
\end{abstract}

\section{Introduction}
Large language models (LLMs) are increasingly applied in reasoning-intensive settings such as coding \citep{jimenez2024swebench}, deductive proof generation \citep{polu2020generative}, planning \citep{valmeekam2023planbench}, multi-hop question answering \citep{yang2018hotpotqa}, and game AI \citep{silver2018general}. Reasoning in LLMs is typically performed by externalizing intermediate tokens through chain-of-thought \citep[CoT;][]{wei2022chain}, and reinforcement learning has been used to further improve reasoning performance \citep{jaech2024openai, guo2025deepseek}. However, even state-of-the-art reasoning models struggle on long-horizon tasks, hitting abrupt performance cliffs at modest depths, and these failures persist even when models are given their full reasoning-token budgets \citep{rameshkumar2025reasoning}. Errors early in a reasoning chain can compound over the remaining steps \citep{dziri2023faith}, forcing models to spend substantial tokens backtracking in order to recover. Even when models avoid such mistakes, reasoning chains can consume a considerable amount of tokens when each reasoning step makes only a small advance towards the solution.

\begin{figure}
\centering
\includegraphics[width=0.95\linewidth]{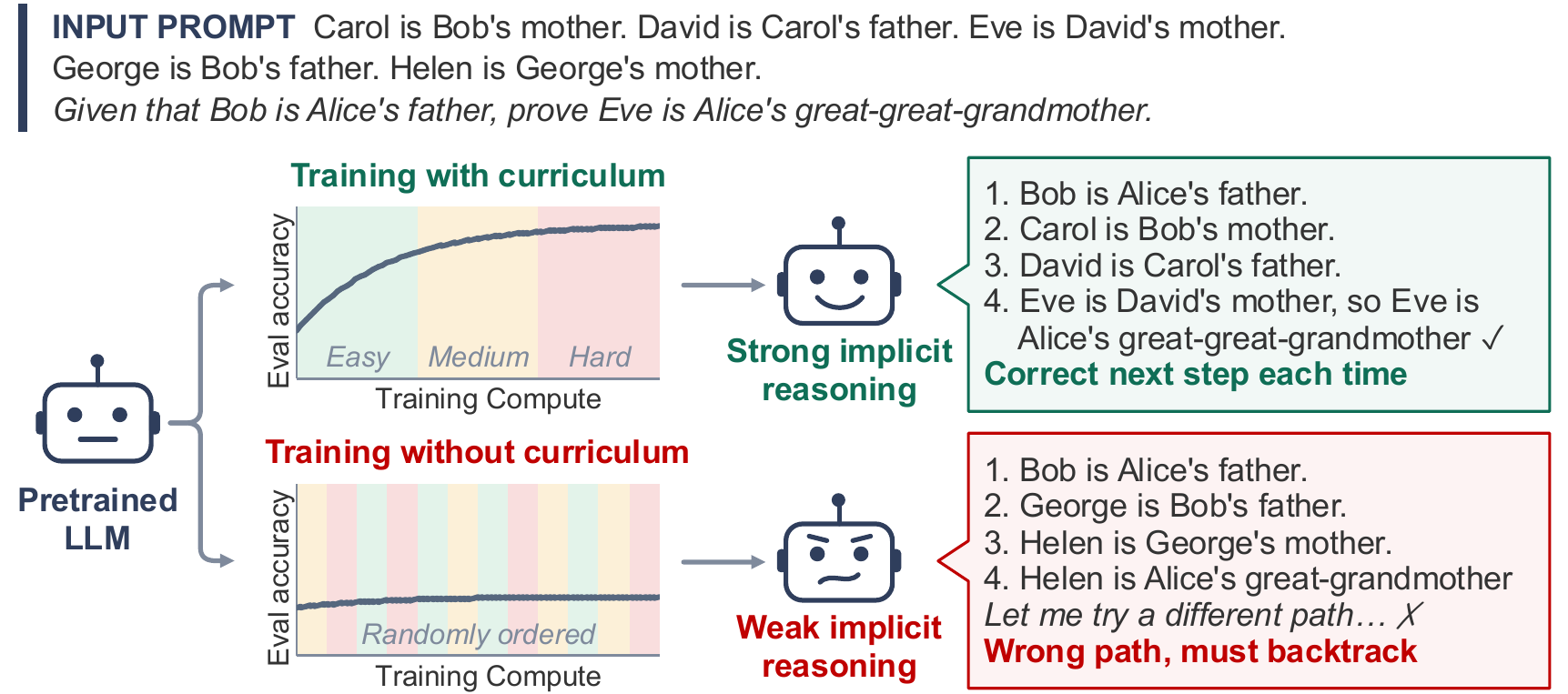}
\caption{Curriculum versus non-curriculum fine-tuning of the same pretrained LLM. The curriculum orders training problems by increasing difficulty, whereas the baseline draws from the full range of difficulties from the start. On the example proof, the curriculum-trained model takes the correct next step each time, while the baseline commits to a wrong branch and must backtrack. Section~\ref{sec:results:curriculum_necessary} reports the measured comparison.}
\label{fig:trace}
\end{figure}

\emph{Implicit reasoning} occurs within a single forward pass and determines how far a model can ``look ahead'' over potential later steps before committing to the next CoT step. This capability becomes increasingly important in long-horizon reasoning tasks, which require the model to reason over greater depths. Consider the example in Figure~\ref{fig:trace}, where a model with weak implicit reasoning commits to a wrong path and must attempt to recover via backtracking, while a model with strong implicit reasoning looks far enough ahead to identify the correct next step directly, shortening the CoT. Despite these benefits, recent work has shown that transformers struggle to learn implicit search beyond shallow depths, with no relief from increased model scale~\citep{saparov2025transformers}. This raises the question: can LLMs be trained to perform long-horizon implicit reasoning, and if so, is there a limit to the depth they can reach?

To address this question, we examine whether the training procedure itself can be shaped to unlock deeper implicit reasoning. Incrementally increasing the depth a model must look ahead during training might offer a natural path to do so, and we therefore investigate whether curriculum learning \citep{bengio2009curriculum} can extend this depth. We generate training data using a modified version of \texttt{DeepRD}~\citep{rameshkumar2025reasoning}, where each example provides a set of premises in natural language over entities and attributes (e.g., in ``Alex is a cat,'' ``Alex'' is an entity and ``cat'' is an attribute) and asks the model to predict the next step of a multi-hop proof. We deliberately keep the curriculum simple, as our goal is not to find the optimal design but to characterize the behavior of curriculum training itself. Our curriculum gradually increases the \emph{lookahead}---the number of steps a model must search ahead to identify the correct next step---jointly with the attribute vocabulary, which we find outperforms a curriculum over lookahead alone (Section~\ref{sec:results:vocab_ablation}). As an illustration, the lookahead in Figure~\ref{fig:trace} is 3, since the model must search 3 steps ahead to conclude that the correct second step is ``Carol is Bob's mother,'' and not ``George is Bob's father.'' By training only on the output of a single forward pass, we isolate implicit reasoning as a pure computational capability, separating it from what can be achieved by externalizing intermediate steps. With this setup, we pose two primary research questions. \textbf{RQ1}: Can LLMs learn to reason implicitly to arbitrary depths, and what training conditions---including curriculum, model size, and initialization---govern this capability? \textbf{RQ2}: Does training a model to perform long-horizon implicit reasoning improve its performance on downstream natural-language reasoning tasks?

We make the following contributions to answer the aforementioned research questions:
\begin{enumerate}[label=\textbf{C\arabic*}, leftmargin=*, itemsep=2pt, topsep=-3pt, parsep=0pt]
    \item We introduce a curriculum over lookahead and attribute vocabulary and show that it extends the depth to which LLMs can reason implicitly. At matched compute, \texttt{Qwen3-0.6B} reaches $98\%$ accuracy on problems up to lookahead $256$ with curriculum vs. $45\%$ without (Section~\ref{sec:results:curriculum_necessary}).
    \item We show that achieved lookahead grows predictably with training compute, following a saturating exponential, suggesting a \emph{maximum intrinsic lookahead} to which each model can learn to reason implicitly (Section~\ref{sec:results:scaling}).
    \item We show that larger models have higher maximum intrinsic lookaheads, that pretrained initialization allows a model to reason implicitly to far greater depths than training from scratch, and that the curriculum step size (the increase in lookahead between successive stages) must be small enough for training to progress. \texttt{Pythia-410M}, \texttt{1.4B}, and \texttt{2.8B} have maximum intrinsic lookaheads of $51$, $82$, and $235$, respectively (Section~\ref{sec:results:model_size}). At a $1.5$M PFLOP budget, \texttt{Pythia-1.4B} achieves lookahead $63$ from pretrained weights vs. $23$ from random initialization (Section~\ref{sec:results:initialization}). With \texttt{Qwen3-0.6B}, step size $64$ stalls at lookahead $128$, while step size $8$ reaches $256$ (Section~\ref{sec:results:step_size}).
    \item We show that the implicit reasoning capability acquired on synthetic data transfers to natural language reasoning. Curriculum-trained models outperform models trained without curriculum at matched compute on $11$ benchmarks, with the largest gains on multi-step deductive reasoning---\texttt{Qwen3-1.7B} gains $14.3$ points on \texttt{ProofWriter} \texttt{OWA} at lookahead $48$ (Section~\ref{sec:results:downstream}).
\end{enumerate}

\section{Related Work}

\paragraph{Reasoning and search in transformers.}
A growing line of work characterizes how LLMs struggle in long-horizon reasoning. \citet{dziri2023faith} argue that transformers reduce multi-step reasoning to pattern matching, so errors compound as tasks grow more complex. Formal analyses of deductive reasoning show that LLMs reason greedily at branch points \citep{saparov2022language} and that their accuracy decreases as the correct proof grows longer \citep{saparov2023testing}. \citet{saparov2025transformers} formalize task complexity via a \emph{lookahead} metric and show that while small transformers can learn to search under a carefully balanced training distribution, learning degrades as graphs grow, with no relief from model scale. \citet{rameshkumar2025reasoning} show that even frontier reasoning models hit a performance cliff at modest lookaheads on graph connectivity and proof planning, without exhausting their token budgets. \citet{heyman2025evaluating} report similar failures on graph-coloring tasks.

\paragraph{Explicit and implicit reasoning.}
Explicit reasoning externalizes tokens via CoT prompting \citep{wei2022chain} and its derivatives \citep{wang2022self, yao2023tree}, extended further by trace-supervised search \citep{lehnert2024beyond, gandhi2024stream} and reinforcement learning \citep{lambert2024tulu, jaech2024openai, guo2025deepseek}. Implicit approaches enhance reasoning without verbalizing intermediate steps via learnable auxiliary tokens \citep{goyal2023think}, soft tokens in place of discrete tokens \citep{hao2024training, shen2025codi, wei2026simcot}, or architectural recurrence \citep{dehghani2018universal, geiping2025scaling}. Mechanistic studies show transformers can reason implicitly within a forward pass \citep{brinkmann2024mechanistic, jenner2024evidence}, and it can emerge through grokking, but only with significant over-training \citep{wang2024grokked}. However, implicit reasoning can reflect shortcut learning rather than generalizable reasoning \citep{lin2025implicit}, and its generalization depends on which compositional structures appear in training \citep{ye2025how}.

\paragraph{Curriculum learning for reasoning.}
In early curriculum learning work, \citet{bengio2009curriculum} train language models on a restricted vocabulary that is progressively expanded; our attribute-vocabulary curriculum applies the same idea to the attributes of synthetic proofs. Curriculum learning has since been used to improve performance on reasoning tasks, for example by gradually removing CoT tokens \citep{deng2024explicit} or replacing them with soft tokens \citep{hao2024training} while keeping the remainder of the prompt invariant. Our curriculum instead never uses CoT and increases the lookahead of the problems. Recent theoretical results show that staged curricula can avoid exponential sample-complexity bottlenecks \citep{wang2025learning, bu2026provable, huang2026transformers} and can enable agents trained with reinforcement learning to perform search on trees that are deeper than those seen during training \citep{yang2026agentic}. Empirically, the picture is mixed: curriculum learning improves learning efficiency \citep{abbe2024far} and mitigates, but does not eliminate, the exponential data cost of implicit $k$-hop reasoning in small from-scratch models \citep{yao2025language}, while difficulty-ordering a fixed dataset gives no robust advantage for explicit CoT post-training \citep{mordig2026rethinking}. \citet{lee2025self} report that pretrained initialization accelerates curriculum progression. Whether a curriculum can unlock implicit reasoning in pretrained LLMs at depths where direct training fails has remained open; our work answers this question.

\section{Methods}

\subsection{Task Formulation and Complexity}
\label{sec:methods:task}

We study whether LLMs can learn to perform long-horizon reasoning implicitly, where the model emits the next proof step directly, without intermediate reasoning tokens. To do so, we focus on the task of predicting the next step in a proof: given a set of premises and a goal conclusion, the model must identify the correct next step in a valid proof of the goal from the premises. A proof is a sequence of proof steps, where each proof step has some premises followed by a conclusion, and the conclusion of one proof step may serve as a premise for a later proof step. Solving the task requires the model to search through possible inference paths, starting from the premises, and find a proof from the premises to the goal.

\paragraph{Training Data.} To generate examples of this task, we use a modified version of the \texttt{DeepRD} dataset. In \texttt{DeepRD}, each problem is generated by first generating a directed acyclic graph (DAG). Each graph is then translated into natural language by mapping every node into an atomic proposition (e.g., ``Alice is a cat'') and every edge into a rule (e.g., ``If Alice is a cat, then Alice is a mammal''). In \texttt{DeepRD}, attribute names, such as ``cat'', are replaced with fictional names (e.g., ``zorbi'') to facilitate the generation of very large reasoning problems and to avoid data contamination issues during evaluation. Since \texttt{DeepRD} is synthetic, we can generate a limitless number of examples for large-scale training. Some examples are shown in Appendix~\ref{app:examples}. A key reason we choose \texttt{DeepRD} is that our curriculum requires a large number of examples whose lookahead is highly controllable and can be made arbitrarily large. Real-world reasoning data cannot meet this requirement, because its complexity is long-tailed~\citep{rameshkumar2025reasoning} and examples with very large lookahead are rare. The representation of each reasoning problem as a graph enables us to precisely control problem complexity: in the graph analogy, finding a proof from a premise to a goal is equivalent to finding a directed path from a start node to a goal node. 

\paragraph{Lookahead.} We measure task complexity using the \emph{lookahead} $\ell$ which is the minimum number of breadth-first search iterations that must be performed from the start node to determine the correct next step in the path to the goal. Intuitively, $\ell$ captures how far the model must search to determine the correct path from incorrect alternatives.\footnote{For a formal definition, see~\citet{rameshkumar2025reasoning, saparov2025transformers}.} This is distinct from the length of the path from the start node to the goal node. For example, if the start node has only one valid child node, no search is needed to determine the next correct step, regardless of the distance of the goal node from the start node. In this case, $\ell = 1$, and lookahead is a more faithful characterization of problem complexity.

\subsection{Curriculum over Lookahead and Attribute Vocabulary}
\label{sec:methods:curriculum}

We aim to test whether curriculum learning can help transformers to learn to perform long-horizon implicit reasoning. With this in mind, we develop a curriculum learning procedure (Figure~\ref{fig:training_loop} and Algorithm~\ref{alg:curriculum}, Appendix~\ref{app:algorithm}). We train the model in stages: at stage $k$, the model is presented with examples whose lookahead is at most $L_k$, where $L_k = k \cdot s$ and $s \geq 1$ is the \emph{curriculum step size}---i.e., the amount by which the lookahead is increased at each stage of the curriculum. For example, with $s = 8$, a 12-stage curriculum advances through $L_k = 8, 16, 24, \ldots, 96$. Figure~\ref{fig:curriculum_diagram} illustrates the first three stages of a curriculum with $s = 1$. At each stage, we perform supervised fine-tuning on examples from \texttt{DeepRD} where, for each example $i$, the lookahead of the example $\ell_i$ is sampled i.i.d.\ from a uniform distribution $\ell_i \sim \text{Uniform}\{1, \ldots, L_k\}$. Including easier examples at each stage helps prevent catastrophic forgetting of previously learned lookaheads. The training objective is cross-entropy on the answer tokens: the tokens identifying the correct next attribute in the proof chain. The model advances from stage $k$ to stage $k+1$ when its exact-match accuracy over the most recent $W$ training examples reaches 98\%.

\paragraph{Attribute-vocabulary curriculum.}
Besides lookahead, the number of distinct attribute labels that appear across training examples also affects the difficulty of the task. We therefore also grow the attribute vocabulary as part of the curriculum, akin to the vocabulary curriculum of \citet{bengio2009curriculum}. The attributes in each generated proof are given fictional labels (e.g., \texttt{zorbi} in ``Alice is zorbi''); the \emph{attribute vocabulary} $\mathcal{V}_k$ is the pool of these attribute labels available to the instances at stage $k$, and $|\mathcal{V}_k|$ denotes its size. We emphasize that $|\mathcal{V}_k|$ does \emph{not} refer to the model's tokenizer vocabulary. The tokenizer and its token set are unchanged, and each attribute label is encoded as ordinary subword tokens. We grow $|\mathcal{V}_k|$ with lookahead $L_k$ so that the vocabulary is always sufficient to provide distinct attributes with distinct labels. Specifically, an instance at lookahead $L_k$ contains at most $2L_k + 1$ distinct attributes: the ``correct chain'' of length $L_k$ plus a distractor chain of length $L_k$ that shares the initial attribute as the correct chain. We set $|\mathcal{V}_k| = 2(L_k + 1)$, which is one more than the upper bound, so every attribute in each instance can receive a distinct label. A pool smaller than $2L_k + 1$ would force two or more attributes to share a label, which would make the problem ill-posed. We explore whether this \emph{joint} curriculum over both lookahead and attribute vocabulary size performs better than a curriculum over lookahead alone (Section~\ref{sec:results:vocab_ablation}). In the latter setting, holding the pool fixed at its maximum size from the start leaves proof structure unchanged and only reduces cross-instance reuse of attribute labels.

\section{Experiments and Results}
\label{sec:experiments}

\paragraph{Models.} We use two model families. \texttt{Qwen3}~\citep[0.6B, 1.7B;][]{yang2025qwen3} gives capable models at low training cost across our many curriculum configurations, and it has fairly strong instruction following capabilities, which facilitates evaluation using downstream benchmarks that require specifically formatted outputs (Section~\ref{sec:results:downstream}). \texttt{Pythia}~\citep[160M, 410M, 1.4B, 2.8B;][]{biderman2023pythia} provides models of different sizes pretrained, where each model is trained on the same data and the same number of tokens, so that they differ mainly in scale, together with base (non-instruction-tuned) checkpoints; our scaling and initialization comparisons (Sections~\ref{sec:results:model_size} and~\ref{sec:results:initialization}) rely on both properties.

\paragraph{Setup.} All training examples are generated with our modified \texttt{DeepRD} generator as described in Sections~\ref{sec:methods:task} and~\ref{sec:methods:curriculum}. Experiments perform full fine-tuning with AdamW, a constant learning rate, and bf16 precision; per-experiment hyperparameters are listed in Appendix~\ref{app:hyperparams}. Training compute is estimated as $C \approx 6ND$~\citep[$N$ parameters, $D$ tokens;][]{kaplan2020scaling}.

\paragraph{Evaluation.} The comparison of training with vs. without curriculum (Section~\ref{sec:results:curriculum_necessary}) uses exact-match accuracy of the predicted next attribute on a held-out set shared by both settings, with lookaheads sampled uniformly from 1 to $L$ (Appendix~\ref{app:eval_methodology}).\footnote{While using curriculum learning, the maximum lookahead of samples reaches $L$ only at the final stage.} All other synthetic-task experiments report \emph{achieved lookahead} $L$, the last curriculum stage completed, i.e., the largest $L$ at which the rolling exact-match training accuracy reaches $98\%$ (Section~\ref{sec:methods:curriculum}).

\paragraph{Baselines.} In both the next-step proof prediction comparisons (Section~\ref{sec:results:curriculum_necessary}) and the downstream transfer experiments (Section~\ref{sec:results:downstream}), the baseline is the same model trained without curriculum, directly on data with a fixed maximum lookahead $L$ from the start, compared against its curriculum-trained counterpart at matched training compute.

\begin{figure}[!htbp]
    \centering
    \vspace{-0.5em}
    \includegraphics[width=0.75\linewidth]{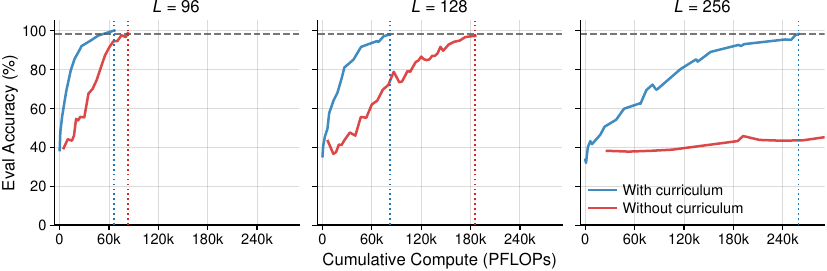}
    \vspace{-0.7em}
    \caption{Eval accuracy of \texttt{Qwen3-0.6B} at maximum $L \in \{96, 128, 256\}$ vs.\ cumulative training compute (PFLOPs). The dashed line marks the $98\%$ threshold; dotted lines mark where each setting crosses it (Table~\ref{tab:curr_vs_direct}).}
\label{fig:curriculum_vs_nocurr_per_L}
\vspace{-0.5em}
\end{figure}

\needspace{4\baselineskip} %
\subsection{Curriculum Learning Enables Deeper Reasoning at Matched Compute}
\label{sec:results:vocab_ablation}
\label{sec:results:curriculum_necessary}

\textbf{Curriculum versus no curriculum.} To determine whether curriculum learning is needed to train models to perform long-horizon reasoning implicitly, we compare training \texttt{Qwen3-0.6B} with curriculum learning at $s=8$ against training without curriculum learning, where the maximum lookahead is fixed from the start. We run this comparison for each maximum lookahead $L \in \{96, 104, 112, 120, 128, 256\}$. Figure~\ref{fig:curriculum_vs_nocurr_per_L} shows evaluation accuracy vs.\ cumulative training compute for $L \in \{96, 128, 256\}$, with the full sweep reported in Appendix~\ref{app:full_curr_vs_nocurr_sweep}. %
\begin{wraptable}[9]{r}{0.5\textwidth}
\vspace{-0.0em} %
\centering
\small
\caption{Cumulative training compute (PFLOPs) to reach $98\%$ accuracy at each target lookahead $L$. A dash marks a setting that never reaches $98\%$ within budget.}
\label{tab:curr_vs_direct}
\vspace{-1.0em} %
\begin{tabular}{crrr}
\toprule
$L$ & Curriculum & No curriculum & Speedup \\
\midrule
96 & 66.5K & 82.8K & $1.25\times$ \\
104 & 66.5K & 106K & $1.59\times$ \\
112 & 74.3K & 140K & $1.89\times$ \\
120 & 74.3K & 106K & $1.42\times$ \\
128 & 81.8K & 186K & $2.27\times$ \\
256 & 259K & -- & -- \\
\bottomrule
\end{tabular}
\end{wraptable}
Table~\ref{tab:curr_vs_direct} summarizes the compute required to reach $98\%$ evaluation accuracy across all six values of $L$. For $L \leq 128$ both methods eventually converge, but training with curriculum is consistently \textbf{$1.25$--$2.27\times$ more compute-efficient}. At maximum lookahead $L{=}256$ the gap becomes qualitative: training without curriculum plateaus at $\sim$45\% accuracy and never converges within the compute budget, while training with curriculum continues climbing to $98\%$.

\begin{wrapfigure}[18]{r}{0.43\textwidth}
\vspace{-1.8em} %
\centering
\includegraphics[width=\linewidth]{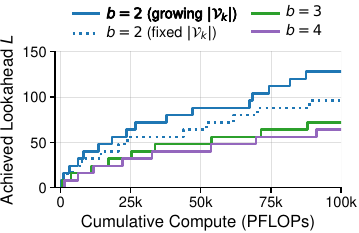}
\captionsetup{skip=4pt} %
\caption{Achieved lookahead $L$ vs.\ cumulative training compute (PFLOPs) for \texttt{Qwen3-0.6B} at $s{=}8$, with the attribute vocabulary grown as $|\mathcal{V}_k| = bL_k + 2$ or held fixed at its maximum size. The bold entry ($b{=}2$, growing $|\mathcal{V}_k|$) is the configuration used in all other experiments.}
\label{fig:fixedV_comparison}
\end{wrapfigure}

\textbf{Joint curriculum over attribute vocabulary.} We next test the design choice introduced in Section~\ref{sec:methods:curriculum}: using a curriculum over both attribute vocabulary size and lookahead, with $|\mathcal{V}_k| = 2(L_k + 1)$, versus using a curriculum over lookahead alone, with the attribute vocabulary fixed at its maximum size throughout training. We train \texttt{Qwen3-0.6B} under both configurations, holding all other hyperparameters constant. Figure~\ref{fig:fixedV_comparison} and Table~\ref{tab:vocab_ablation_compute} show that the joint curriculum completes $L{=}128$ within $100$K PFLOPs, where the fixed-vocabulary configuration has completed only $L{=}96$. One plausible explanation is the following: at a fixed (large) attribute vocabulary, the model sees fewer examples of each attribute label in diverse reasoning contexts, encouraging memorization rather than generalization, while the joint curriculum keeps the attribute vocabulary small in early stages and grows it progressively as the model masters larger lookaheads. We therefore conduct all subsequent experiments using the joint curriculum over lookahead and attribute vocabulary size.

\textbf{Vocabulary growth rate.} Beyond whether to grow the attribute vocabulary at all, we also test how quickly it should grow. We compare $|\mathcal{V}_k| = bL_k + 2$ for $b \in \{2, 3, 4\}$ on \texttt{Qwen3-0.6B} with $s=8$ under an otherwise identical curriculum (Figure~\ref{fig:fixedV_comparison}). At $b = 2$, this recovers our original schedule, $|\mathcal{V}_k| = 2L_k + 2 = 2(L_k + 1)$. For a given lookahead, a larger $b$ ``widens'' the deepest instances from $2$ branches to $b$, rather than simply enlarging the attribute vocabulary while holding proof structure fixed, since $|\mathcal{V}_k|$ is one more than the maximum number of attributes in an instance. This comparison therefore varies proof \emph{breadth} alongside attribute vocabulary size. Within a budget of 100K PFLOPs, the three settings achieve $L{=}128$, $72$, and $64$, respectively. Thus, the curriculum design that advances fastest is the one that grows the attribute vocabulary while holding the deepest instances at two branches ($b=2$), and we use this configuration in all subsequent experiments.

\subsection{Achieved Lookahead Grows Predictably toward an Apparent Ceiling}
\label{sec:results:scaling}

\begin{figure}[]
    \centering
    \includegraphics[width=0.8\linewidth]{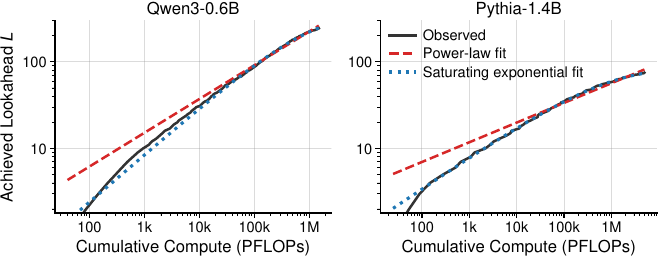}
    \vspace{-0.6em}
\caption{Achieved lookahead $L$ vs.\ cumulative training compute (PFLOPs, log--log scale) for \texttt{Qwen3-0.6B} (\emph{left}) and \texttt{Pythia-1.4B} (\emph{right}) trained under the curriculum with $s{=}1$, with the power-law and saturating exponential fits overlaid on the observed trajectory.}
    \label{fig:scaling_powerlaw}
\end{figure}

Having shown that curriculum is instrumental in teaching models to perform long-horizon implicit reasoning, we examine how achieved lookahead grows with compute. Figure~\ref{fig:scaling_powerlaw} plots achieved $L$ against cumulative training compute (PFLOPs, log--log) for the trajectories of two models trained under our curriculum with $s=1$: \texttt{Qwen3-0.6B} and \texttt{Pythia-1.4B}. In both cases achieved $L$ grows smoothly with compute but decelerates, curving below a straight line on log--log axes. We fit two functional forms\footnote{We also fit a shifted power law, $L = a(C + C_s)^p$, and a saturating power law, $L = L_\infty - bC^{-k}$. The fitted shift $C_s$ is close to zero, so the fitted function is effectively equivalent to a plain power law. The fitted exponent $k$ is close to zero, so the fitted function is effectively equivalent to a logarithmic curve. The logarithmic
form produces a poorer fit than the saturating exponential on every trajectory.}: a power law $L = aC^{p} + L_0$, the simplest unbounded form, and a saturating exponential $L = L_{\max}\bigl(1 - \exp\bigl(-(C/C_0)^{k}\bigr)\bigr)$, the simplest bounded form, with a ceiling $L_{\max}$.

Under the constraint $L_0 \geq 0$ (to ensure positive lookahead) the fitted offset vanishes, leaving a pure power law. Table~\ref{tab:fit_forms} gives the fitted forms and their in-sample RMSE.

\begin{table}[!htbp]
\centering
\small
\vspace{-0.4em}
\caption{Fitted forms of achieved lookahead $L$ against cumulative training compute $C$ (PFLOPs), with in-sample RMSE.}
\label{tab:fit_forms}
\vspace{-0.9em}
\begin{tabular}{@{}lllr@{}}
\toprule
Trajectory & Form & Fitted equation & RMSE \\
\midrule
\multirow{2}{*}{\texttt{Qwen3-0.6B}} & Power law & $L \approx 1.03\,C^{0.390}$ & 6.10 \\
 & Saturating exponential & $L \approx 324\bigl(1 - \exp\bigl(-(C/7.7{\times}10^{5})^{0.547}\bigr)\bigr)$ & 2.22 \\
\midrule
\multirow{2}{*}{\texttt{Pythia-1.4B}} & Power law & $L \approx 2.45\,C^{0.228}$ & 2.98 \\
 & Saturating exponential & $L \approx 82\bigl(1 - \exp\bigl(-(C/5.0{\times}10^{5})^{0.371}\bigr)\bigr)$ & 0.53 \\
\bottomrule
\end{tabular}
\vspace{-0.4em}
\end{table}

The saturating exponential achieves lower in-sample RMSE than the power law by factors of $2.7$ and $5.6$ on \texttt{Qwen3-0.6B} and \texttt{Pythia-1.4B}, respectively.

\begin{wraptable}[9]{r}{0.5\textwidth}
\vspace{-1.2em} %
\centering
\small
\setlength{\tabcolsep}{4pt}
\caption{Held-out RMSE for achieved lookahead $L$. $p$-value: significance of the saturating exponential's advantage under the per-stage iteration test, two-sided. Additional fits on less-trained trajectories in Table~\ref{tab:curve_heldout_all}.}
\vspace{-0.5em}
\label{tab:curve_heldout}
\vspace{-4pt} %
\begin{tabular}{lrrr}
\toprule
& \multicolumn{2}{c}{Held-out RMSE} & \\
\cmidrule(lr){2-3}
Trajectory & Power law & Sat.\ exp. & $p$-value \\
\midrule
\texttt{Qwen3-0.6B} & 23.88 & 17.08 & $7{\times}10^{-10}$ \\
\texttt{Pythia-1.4B} & 16.14 & 0.68 & $1.5{\times}10^{-4}$ \\
\bottomrule
\end{tabular}
\end{wraptable}

We next test whether this in-sample advantage extends to prediction. We fit both forms on the first $70\%$ of each trajectory's stage completions, ordered by cumulative compute, and evaluate on the remaining $30\%$. Held-out RMSE is $1.4\times$ and $23.8\times$ lower for the saturating exponential on \texttt{Qwen3-0.6B} and \texttt{Pythia-1.4B} (Table~\ref{tab:curve_heldout}). A Diebold--Mariano test \citep{diebold1995comparing} is a standard comparison of held-out forecast errors. Because errors along a single trajectory are serially correlated, however, we instead apply the test to the number of training iterations required to complete each stage, whose residuals are approximately independent (Appendix~\ref{app:increment-test}). On both trajectories, the saturating exponential predicts the held-out duration of each stage significantly better than the power law ($p = 7{\times}10^{-10}$ for \texttt{Qwen3-0.6B} and $p = 1.5{\times}10^{-4}$ for \texttt{Pythia-1.4B}; Table~\ref{tab:curve_heldout}). The advantage also increases with lookahead, as the gap between the two forms' held-out prediction errors widens at later stages. Appendix~\ref{app:curve-fit} reports the results of the same comparisons on \texttt{Pythia-410M} and \texttt{Pythia-2.8B}.

Within our observed range both forms describe the data closely, but achieved $L$ decelerates faster than a power law allows. The saturating fit is better, and it assigns each model a finite ceiling $L_{\max}$. The slowdown is particularly pronounced for \texttt{Pythia-1.4B}. Within a budget of 5M PFLOPs, it reaches $L{=}75$, $91\%$ of the fitted asymptote $L_{\max}\approx82$. Its last five completed stages, $L{=}71$--$75$, consume $2.10$M PFLOPs, approximately $42\%$ of the budget, and the per-stage cost is still increasing at the end of training (Appendix~\ref{app:per_stage_compute}, Figure~\ref{fig:per_stage_cost_pythia14b}).

These trajectories suggest a model- and procedure-specific ceiling on achievable lookahead: for a given model, our curriculum appears to drive $L$ upward with diminishing returns toward that ceiling.

\subsection{Small Step Sizes Are Required for Deep Curriculum Progression}
\label{sec:results:step_size}%
\begin{wrapfigure}[14]{l}{0.43\textwidth}
\vspace{-0.9em}
\centering
\includegraphics[width=\linewidth]{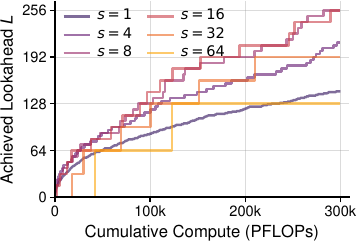}
\vspace{-1.1em}
\captionsetup{skip=4pt} %
\caption{Achieved lookahead $L$ vs.\ cumulative training compute for \texttt{Qwen3-0.6B} with step sizes $s \in \{1, 4, 8, 16, 32, 64\}$.}
\label{fig:stepsize_sweep}
\end{wrapfigure} %

Our curriculum has one main hyperparameter, the step size $s$, which controls the increase in maximum lookahead $L$ between consecutive stages. We examine how $s$ affects the lookahead the curriculum reaches within a fixed compute budget. We tune \texttt{Qwen3-0.6B} using our curriculum at step sizes $s \in \{1, 4, 8, 16, 32, 64\}$, each with a target lookahead of $L{=}256$.

Figure~\ref{fig:stepsize_sweep} shows that the step size must be neither too large nor too small. A step size that is too large stalls curriculum advancement: $s{=}64$ stalls at $L{=}128$ and $s{=}32$ at $L{=}192$. A step size that is too small does not stall but advances slowly: within 300K PFLOPs, $s{=}1$ reaches only $L{=}145$ and $s{=}4$ reaches $L{=}212$, whereas $s{=}8$ and $s{=}16$ achieves the target of $L{=}256$. A plausible explanation for stalling is that larger $s$ makes training more similar to the no-curriculum case: at $s$ equal to the target lookahead the curriculum has a single stage, which is exactly the no-curriculum case, and Section~\ref{sec:results:curriculum_necessary} shows that this stalls at $L{=}256$. The higher the target, the smaller $s$ must be to avoid stalling.

\subsection{Maximum Intrinsic Lookahead Grows with Model Scale}
\label{sec:results:model_size}

\begin{wrapfigure}[15]{r}{0.43\textwidth}
\vspace{-1.0em} %
\centering
\includegraphics[width=\linewidth]{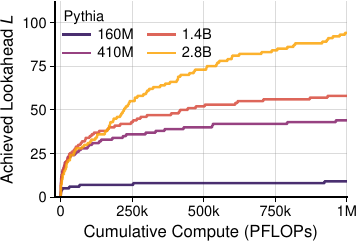}
\captionsetup{skip=4pt} %
\vspace{-1.4em}
\caption{Achieved lookahead $L$ vs.\ cumulative training compute for the four \texttt{Pythia} sizes at $s{=}1$. At 1M PFLOPs, the curves reach $L{=}9$, $44$, $58$, and $94$, from smallest to largest.}
\label{fig:family_scaling}
\end{wrapfigure}
Prior work finds that scaling model size alone does not improve implicit reasoning~\citep{saparov2025transformers}. We test whether this conclusion holds with curriculum learning, and whether the curriculum's progression rate depends on model size. We train \texttt{Pythia} models of four different sizes (160M, 410M, 1.4B, 2.8B) using our curriculum with $s = 1$, with learning rate and batch size individually tuned per model size (Appendix~\ref{app:hyperparams}). 

At matched compute, larger models reach greater lookaheads once the training budget is sufficiently large (Figure~\ref{fig:family_scaling}). \texttt{Pythia-2.8B} achieves the deepest lookahead among the tested \texttt{Pythia} models ($L{=}133$). It completes $L{=}130$ using 2.40M PFLOPs, less than half the compute \texttt{Pythia-1.4B} uses to reach $L{=}75$ (4.97M PFLOPs). \texttt{Pythia-160M} plateaus at $L{=}9$ over the observed trajectory, suggesting that the model's capacity is insufficient for curriculum progression.

\paragraph{Scaling of the fitted asymptote with model size.}
The saturating exponential's asymptote $L_{\max}$ is itself model-dependent. We fit the saturating-exponential curve introduced in Section~\ref{sec:results:scaling} separately to each \texttt{Pythia} trajectory except that of \texttt{Pythia-160M}, which plateaus at $L{=}9$.
The fitted asymptote rises with model size: $L_{\max} \approx 51$, $82$, and $235$ for \texttt{Pythia-410M}, \texttt{-1.4B}, and \texttt{-2.8B} respectively (per-size fits in Appendix~\ref{app:curve-fit}, Figure~\ref{fig:lmax_scaling}; fitted parameters in Table~\ref{tab:fit-eqns}). Parameter count is not the whole story, however: \texttt{Qwen3-0.6B} has a fitted asymptote of $L_{\max}{\approx}324$, well above the larger \texttt{Pythia-1.4B}'s ${\approx}82$, suggesting that factors beyond parameter count also affect a model's maximum intrinsic lookahead. Bootstrap intervals (Table~\ref{tab:fit-eqns}) show that the ordering of these ceilings is robust.

\subsection{Pretrained Initialization Reaches Greater Lookahead at Matched Compute}
\label{sec:results:initialization}
\begin{figure}[!htbp]
\centering
\includegraphics[width=0.9\linewidth]{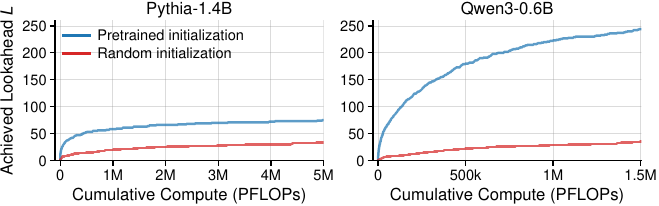}
\caption{Achieved lookahead $L$ vs.\ cumulative training compute (PFLOPs), starting from the pretrained checkpoint vs.\ random initialization. \emph{Left:} \texttt{Pythia-1.4B}, \emph{Right:} \texttt{Qwen3-0.6B}.}
\label{fig:init}
\end{figure}

\citet{lee2025self} report that pretrained initialization accelerates curriculum progression in arithmetic and maze-solving tasks. We test whether this effect holds for our curriculum learning in reasoning and quantify the size of the gap. 
We train \texttt{Pythia-1.4B} and \texttt{Qwen3-0.6B}, each twice under our curriculum with $s{=}1$: once from the released pretrained checkpoint (full fine-tuning) and once from random weights (training from scratch). Learning rate and batch size are tuned separately for each condition, since random initialization makes little progress under the small learning rate suited to fine-tuning (Appendix~\ref{app:hyperparams}).
Pretraining gives a large head start in both families (Figure~\ref{fig:init}). Within a budget of 1.5M~PFLOPs, random initialization reaches $L{=}23$ for \texttt{Pythia-1.4B} and $L{=}35$ for \texttt{Qwen3-0.6B}. Pretrained initialization reaches the same lookaheads with ${\sim}53\times$ and ${\sim}117\times$ less compute, respectively (Table~\ref{tab:init_compute}, Appendix~\ref{app:init_table}).
That the stronger, more heavily pretrained model benefits more suggests the curriculum builds on capabilities already present in the pretrained weights rather than instilling them from scratch. Applying our curriculum to \texttt{Pythia-1.4B} checkpoints saved after different amounts of pretraining shows the same pattern within a single model, as checkpoints with more pretraining need less compute to advance through the curriculum (Appendix~\ref{app:init_table}).

\paragraph{Pretrained models have a larger fitted asymptote.} We train the randomly initialized \texttt{Qwen3-0.6B} long enough to fit the curve of Section~\ref{sec:results:scaling}, which gives $L_{\max}\approx120$, against $\approx324$ from pretrained initialization. This suggests that without pretrained knowledge, the model not only progresses more slowly but also reaches a lower ceiling.

\subsection{Implicit Reasoning Learned on Synthetic Data Transfers Downstream}
\label{sec:results:downstream}

We now ask whether the implicit-reasoning capability acquired through curriculum learning on synthetic data transfers to natural language reasoning benchmarks. We fine-tune \texttt{Qwen3-1.7B} with our curriculum ($s=1$). To avoid catastrophic forgetting of the instruction-following capability that downstream benchmarks require, we draw 6\% of training examples, outside the curriculum, from the instruction-tuning corpus \texttt{Dolci-Instruct-SFT}~\citep{olmo3_2025}.

We evaluate at curriculum stages $L \in \{8, 16, 32, 48\}$ on 14 downstream reasoning tasks spanning logical, multi-hop, constraint-satisfaction, and graph reasoning~\citep{tafjord2021proofwriter, clark2020ruletaker, lin2025zebralogic, saparov2023testing, han2024folio, chen2025justlogiccomprehensivebenchmarkevaluating, ICLR2025_862819c2, sinha2019clutrr, parmar2024logicbench, suzgun-etal-2023-challenging, liu2020logiqa, zhong2023agieval, wang2023nlgraph} (full list in Table~\ref{tab:downstream_curriculum_small}; descriptions in Appendix~\ref{app:benchmark_descriptions}). Evaluating at multiple stages tracks how the transfer evolves as the curriculum deepens. We compare against the same model trained without curriculum, directly at $L{=}48$, compute-matched to the final curriculum checkpoint.

Table~\ref{tab:downstream_curriculum_small} reports the results, and Appendix~\ref{app:downstream_ci} additionally reports the base model and an instruction-only control. At $L{=}48$, training with curriculum outperforms training without curriculum on 11 of the 14 benchmarks, with statistically significant gains on \texttt{ProofWriter} (both settings), \texttt{JustLogic}, \texttt{ZebraLogic-MC}, \texttt{PrOntoQA}, \texttt{ParaRules}, \texttt{FOLIO}, and \texttt{LogicBench-BQA}. On the three benchmarks where the model trained without curriculum scores higher, none of the differences is significant. These results suggest that training a model with curriculum learning to perform long-horizon implicit reasoning can improve its downstream reasoning. Optimizing the training procedure to maximize these gains is left to future work.

\begin{wraptable}[21]{r}{0.5\textwidth}
\centering
\small
\vspace{-1.5em}
\caption{Downstream reasoning accuracy (\%) for \texttt{Qwen3-1.7B} at curriculum stages $L$, vs.\ trained directly at $L{=}48$ without curriculum at matched compute. $\Delta$: accuracy of curriculum at $L{=}48$ minus that of no curriculum. Best per row in bold. $^{*}$: curriculum wins are statistically significant (two-sided exact McNemar test on paired items, $p<0.05$); confidence intervals in Appendix~\ref{app:downstream_ci}. $^\dagger$Extended by us to test for greater depths (Appendix~\ref{app:benchmark_descriptions}).}
\label{tab:downstream_curriculum_small}
\vspace{-1em} %
\footnotesize
\setlength{\tabcolsep}{2.5pt}
\renewcommand{\arraystretch}{0.9}
\resizebox{\linewidth}{!}{%
\begin{tabular}{@{}l llll l r@{}}
\toprule
& \multicolumn{4}{c}{Curriculum $L$} & \multicolumn{1}{c}{\multirow{2}{*}[-2pt]{\shortstack{\textbf{No}\\\textbf{curr.}}}} & \multicolumn{1}{r@{}}{\multirow{2}{*}[-2.2pt]{\ooalign{\hphantom{$+13.5$}\cr\hfil$\Delta$\hfil\cr}}} \\
\cmidrule(lr){2-5}
\textbf{Benchmark} & \textbf{8} & \textbf{16} & \textbf{32} & \textbf{48} & & \\
\midrule
ProofWriter OWA & 41.1 & 33.6 & 39.3 & \textbf{54.9}$^{*}$ & 40.6 & $+14.3$ \\
ProofWriter CWA & 66.8 & 57.2 & 67.6 & \textbf{67.9}$^{*}$ & 55.2 & $+12.8$ \\
ParaRules & 46.1 & 36.2 & 39.1 & \textbf{55.1}$^{*}$ & 50.4 & $+4.7$ \\
ZebraLogic-MC & 47.4 & 53.4 & 51.8 & \textbf{53.5}$^{*}$ & 48.3 & $+5.2$ \\
PrOntoQA$^\dagger$ & 52.5 & 55.3 & 56.8 & \textbf{57.1}$^{*}$ & 52.0 & $+5.1$ \\
FOLIO & 45.8 & \textbf{46.3} & 45.8 & 45.8$^{*}$ & 41.4 & $+4.4$ \\
JustLogic & 51.1 & 52.6 & 45.5 & \textbf{53.5}$^{*}$ & 46.1 & $+7.4$ \\
ProverQA & 42.0 & 44.4 & 46.3 & 47.8 & \textbf{49.2} & $-1.4$ \\
CLUTRR & 25.5 & 27.2 & 23.2 & \textbf{28.6} & 26.7 & $+1.9$ \\
LogicBench-BQA & \textbf{74.2} & 72.5 & 74.1 & 73.6$^{*}$ & 71.2 & $+2.4$ \\
RuleTaker & 60.5 & 60.6 & \textbf{63.5} & 60.0 & 58.8 & $+1.2$ \\
Web of Lies$^\dagger$ & 55.5 & 56.5 & 57.2 & \textbf{57.9} & 57.8 & $+0.2$ \\
LogiQA & 42.6 & \textbf{44.9} & \textbf{44.9} & 42.4 & 43.4 & $-0.9$ \\
NLGraph & 41.5 & 43.5 & 44.3 & 43.7 & \textbf{46.2} & $-2.5$ \\
\bottomrule
\end{tabular}}
\end{wraptable}

\paragraph{General model capability is preserved.} On eight language-understanding benchmarks and \texttt{IFEval}, the curriculum-trained \texttt{Qwen3-1.7B} exceeds the non-finetuned model at every curriculum stage (eight-benchmark mean ranging from 63.4 to 64.2 for the various curriculum approaches vs.\ 61.7 for non-finetuned model) and stays within one point of a compute-matched model fine-tuned only on instruction data (Table~\ref{tab:general_capability}, Appendix~\ref{app:general_capability}). Thus, we find no evidence that gains in implicit reasoning come at the expense of general capability.

\section{Conclusion}
\label{sec:conclusion}

We investigate whether LLMs can be trained to perform implicit reasoning at greater depths using a curriculum over lookahead and attribute vocabulary size that gradually increases the maximum number of steps the model must search ahead to identify the next step in a proof. We find that curriculum learning enables models to reach greater depths than no-curriculum baselines. Achieved lookahead grows predictably with training compute but decelerates and fine-tuning progresses, and across both model families, the trajectories are better described by a saturating exponential than by an unbounded power law, suggesting a maximum intrinsic lookahead that varies across models. Our fits suggest that model size and pretrained initialization raise the maximum intrinsic lookahead, while step sizes that are too large cause the curriculum to stall. Finally, the capability acquired on synthetic data can transfer to downstream natural-language reasoning benchmarks, showing that curriculum-trained implicit reasoning can generalize beyond the synthetic training distribution.

Future work includes separating the effects of architecture and pretraining on the fitted asymptotes, applying the curriculum to semantically richer generators with a controllable difficulty parameter, and training curriculum-trained models with reinforcement learning to measure the effect on chain-of-thought length and backtracking.

\subsubsection*{Reproducibility Statement}

All required details to reproduce our results are contained in the paper and its
appendices. Additionally, our full code is freely available at \url{https://github.com/MaxNickell/implicit-reasoning-curriculum}. The task formulation and the curriculum over lookahead and attribute
vocabulary are specified in Sections~\ref{sec:methods:task}
and~\ref{sec:methods:curriculum}; the training procedure is given as pseudocode in
Appendix~\ref{app:algorithm}, and worked examples of generated training instances
appear in Appendix~\ref{app:examples}. Training data is produced with a modified
version of the \texttt{DeepRD} generator, with our
modifications described in Section~\ref{sec:methods:task}. Model configurations,
optimizer settings, batch sizes, and learning rates for every
experiment are listed in Appendix~\ref{app:hyperparams}. The downstream evaluation
protocol, including benchmarks, prompting format, and scoring, is given in
Appendices~\ref{app:eval_methodology} and~\ref{app:benchmark_descriptions}; all
models are evaluated under an identical protocol. Confidence intervals and the significance tests behind our
downstream comparisons are reported in Appendix~\ref{app:downstream_ci}, and the
curve-fitting procedure behind our $L(C)$ claims, including held-out validation, is
detailed in Appendix~\ref{app:curve-fit}. We note that every downstream number comes
from a single training seed, and that the intervals we report reflect evaluation
sampling only.



\subsubsection*{Acknowledgments}
This research was supported by computational resources from Purdue University's Rosen Center for Advanced Computing and Contextual AI, Inc.

\bibliographystyle{iclr2027_conference}
\bibliography{main}

\clearpage %
\appendix
\section{Curriculum Training Algorithm}
\label{app:algorithm}

The pseudocode below specifies the curriculum training procedure described in Section~\ref{sec:methods:curriculum}. Stages advance based on a rolling exact-match accuracy criterion over the $W$ most recent training examples, after a per-stage minimum step floor (200 steps, see Appendix~\ref{app:hyperparams}) is reached.

\begin{figure}[H]
\centering
\includegraphics[width=0.85\linewidth]{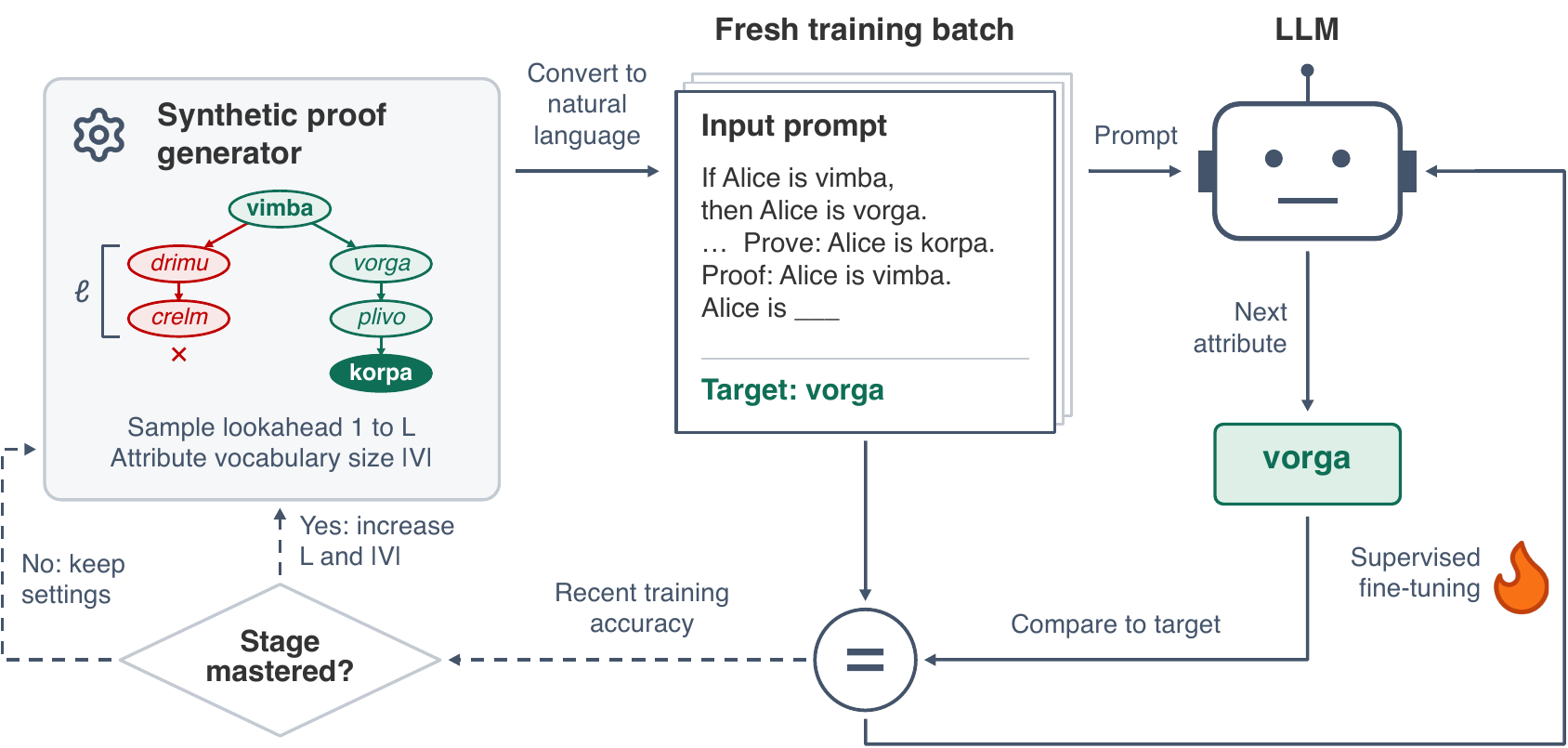}
\caption{Training loop of our curriculum over lookahead and attribute vocabulary, as described in Section~\ref{sec:methods:curriculum}.}
\label{fig:training_loop}
\end{figure}

\begin{algorithm}[H]
\caption{Curriculum Training over Lookahead and Attribute Vocabulary}
\label{alg:curriculum}
\begin{algorithmic}[1]
\Require Model $f_\theta$ with parameters $\theta$, step size $s$, number of stages $K$, threshold $\tau$, window $W$, batch size $B$, minimum steps per stage $M$
\State Initialize $\theta$; set $k \leftarrow 1$
\While{$k \leq K$}
  \State $L_k \leftarrow k \cdot s$
  \State $|\mathcal{V}_k| \leftarrow 2(L_k + 1)$
  \State Sample $\ell_i \sim \mathrm{Uniform}\{1,\dots,L_k\}$ independently for $i = 1, \dots, B$
  \State Generate batch $\{(x_i, y_i)\}_{i=1}^{B}$ with lookaheads $\ell_i$ and attribute labels drawn from a vocabulary of size $|\mathcal{V}_k|$, where $x_i$ is the premises and query and $y_i$ is the next proof step
  \State $\mathcal{L} \leftarrow \frac{1}{B}\sum_{i=1}^{B} \text{CrossEntropy}(f_\theta(x_i), y_i)$ on answer tokens
  \State Update $\theta$ to minimize $\mathcal{L}$
  \If{at least $M$ steps have been taken in stage $k$ \textbf{and} exact-match accuracy over the most recent $W$ examples $\geq \tau$}
    \State $k \leftarrow k + 1$
  \EndIf
\EndWhile
\State \Return $\theta$
\end{algorithmic}
\end{algorithm}

\section{Experimental Details}
\label{app:hyperparams}

We list the training hyperparameters for each experiment in Section~\ref{sec:experiments}. All experiments use AdamW ($\beta_1=0.9$, $\beta_2=0.999$, $\epsilon=10^{-8}$) with a constant learning rate, no weight decay, gradient clipping at $1.0$, and bf16 mixed precision; we find scheduled learning rates consistently degrade or fail to improve over constant LR. All experiments use a fixed random seed of $1234$. Unless noted, every experiment uses 100 warmup steps, $\tau = 0.98$ rolling-accuracy advancement threshold over a rolling window of $W=800$ examples, and a 200-step floor per stage.
Training uses predominantly 2--4$\times$ NVIDIA H100 80GB GPUs, with 8$\times$ NVIDIA H200 used for \texttt{Pythia-2.8B} and 6$\times$ NVIDIA A100 80GB used for \texttt{Pythia-160M} and parts of the \texttt{Qwen3-1.7B} downstream training. Gradient accumulation is chosen to match the effective batch size listed per experiment. Downstream evaluation uses a single NVIDIA A100 GPU.

\paragraph{Notation.} ``Batch size'' refers to the combined effective batch size across GPUs. All \texttt{Pythia} and \texttt{Qwen3} models are initialized from the released checkpoints (\texttt{Pythia}: base; \texttt{Qwen3}: the post-trained release) unless explicitly marked random init.

\subsection{Joint Curriculum vs.\ Fixed-Vocabulary (Section~\ref{sec:results:vocab_ablation})}

Both configurations use \texttt{Qwen3-0.6B} (full FT) with the same optimization stack as Section~\ref{sec:results:curriculum_necessary}: batch~768, LR $5\times 10^{-5}$, curriculum step $s=8$ to $L{=}256$. They differ only in attribute-vocabulary handling: the joint configuration scales the attribute vocabulary size with stage as $|\mathcal{V}_k| = 2(L_k + 1)$, while the fixed-vocabulary configuration holds $|\mathcal{V}|$ at the maximum from the start. The runs with $|\mathcal{V}_k| = bL_k + 2$ for $b \in \{3, 4\}$ use the same settings.

\subsection{Curriculum vs.\ No-Curriculum (Section~\ref{sec:results:curriculum_necessary})}

\begin{table}[H]
\centering
\small
\begin{tabular}{lcc}
\toprule
Setting & Curriculum & No-curriculum \\
\midrule
Model & \texttt{Qwen3-0.6B} (full FT) & \texttt{Qwen3-0.6B} (full FT) \\
Batch size & 768 & 768 \\
Learning rate & $5 \times 10^{-5}$ & $5 \times 10^{-5}$ \\
Curriculum step $s$ & 8 & N/A \\
$L$ & $8\rightarrow256$ & $\{96, 104, 112, 120, 128, 256\}$ \\
\bottomrule
\end{tabular}
\caption{Hyperparameters for Section~\ref{sec:results:curriculum_necessary} (\texttt{Qwen3-0.6B}).}
\end{table}

\subsection{Scaling Behavior (Section~\ref{sec:results:scaling})}

\texttt{Qwen3-0.6B} uses batch 768 and LR $5\times 10^{-5}$ with $s=1$ and a target of $L{=}256$; \texttt{Pythia-1.4B} uses the per-model settings in Table~\ref{tab:modelsize_hp} with $s=1$ and a target of $L{=}96$.

\subsection{Curriculum Step Size Sweep (Section~\ref{sec:results:step_size})}

All six step sizes share the same \texttt{Qwen3-0.6B} optimization stack: batch 768, LR $5\times10^{-5}$, curriculum to $L{=}256$.

\subsection{Model Size Sweep (Section~\ref{sec:results:model_size})}
\label{app:hp_modelsize}

Batch size and learning rate are tuned per model size by a small grid search; for each model, we select the (batch size, learning rate) pair that produces the fastest curriculum advancement during early stages.

\begin{table}[H]
\centering
\small
\begin{tabular}{lcc}
\toprule
Model & Batch size & LR \\
\midrule
\texttt{Pythia-160M} & 192 & $1\times10^{-4}$ \\
\texttt{Pythia-410M} & 192     & $5\times10^{-5}$   \\
\texttt{Pythia-1.4B} & 192 & $2.7\times10^{-5}$ \\
\texttt{Pythia-2.8B} & 1{,}536 & $2.7\times10^{-5}$ \\
\bottomrule
\end{tabular}
\caption{Per-model hyperparameters for Section~\ref{sec:results:model_size} (\texttt{Pythia} model-size sweep). All sizes: $s{=}1$, AdamW, constant LR.}
\label{tab:modelsize_hp}
\end{table}

\subsection{Initialization (Section~\ref{sec:results:initialization})}

Random initialization makes little progress under the small learning rate suited to fine-tuning, so each condition uses its own best-found learning rate and batch size.

\begin{table}[H]
\centering
\small
\begin{tabular}{lccc}
\toprule
Condition & Batch size & LR & Init \\
\midrule
\texttt{Pythia-1.4B}, pretrained & 192 & $2.7\times10^{-5}$ & \texttt{Pythia-1.4B} base \\
\texttt{Pythia-1.4B}, random init & 768 & $1\times10^{-4}$ & default initializer \\
\texttt{Qwen3-0.6B}, pretrained & 768 & $5\times10^{-5}$ & \texttt{Qwen3-0.6B} post-trained release \\
\texttt{Qwen3-0.6B}, random init & 1{,}536 & $3\times10^{-4}$ & default initializer \\
\bottomrule
\end{tabular}
\caption{Hyperparameters for Section~\ref{sec:results:initialization} ($s{=}1$; curriculum to $L{=}96$ for \texttt{Pythia-1.4B} and to $L{=}256$ for \texttt{Qwen3-0.6B}). The \texttt{Qwen3-0.6B} pretrained row reuses the training from Section~\ref{sec:results:scaling}.}
\end{table}

\subsection{Downstream Reasoning Transfer (Section~\ref{sec:results:downstream})}

\texttt{Qwen3-1.7B} uses a 6\% \texttt{Dolci-Instruct-SFT} mix-in alongside the graph-search curriculum; the matched-compute baseline trains the same model on the same mix without curriculum, directly at $L{=}48$. The instruction-only control trains the same model with the same optimizer settings on 100\% \texttt{Dolci-Instruct-SFT}, with no graph-search data. All three configurations wrap inputs in a chat template with \texttt{enable\_thinking=False}.

\begin{table}[H]
\centering
\small
\begin{tabular}{lc}
\toprule
Configuration & Mix \\
\midrule
\texttt{Qwen3-1.7B}, curriculum & 6\% \\
\texttt{Qwen3-1.7B}, no curriculum (direct at $L{=}48$) & 6\% \\
\texttt{Qwen3-1.7B}, instruction only & 100\% \\
\bottomrule
\end{tabular}
\caption{Hyperparameters for Section~\ref{sec:results:downstream}. All: AdamW, constant LR $5\times10^{-5}$, batch 768, chat template with \texttt{enable\_thinking=False}. The curriculum uses $s{=}1$ up to $L{=}48$.}
\end{table}

\section{Detailed Experimental Results}
\label{app:detailed_results}

This appendix tabulates the core experiments of Section~\ref{sec:experiments}, with one subsection per experiment.

\subsection{Full Per-$L$ Curriculum vs.\ No-Curriculum Sweep (Section~\ref{sec:results:curriculum_necessary})}
\label{app:full_curr_vs_nocurr_sweep}

\begin{figure}[t]
\centering
\includegraphics[width=\linewidth]{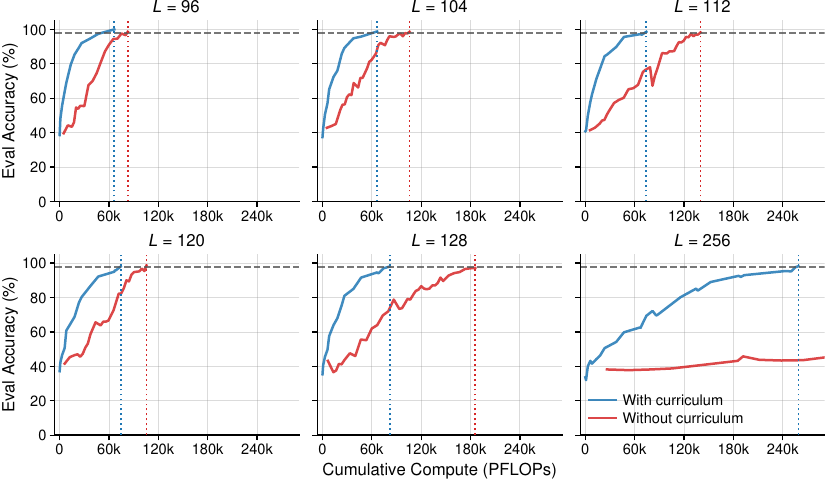}
\caption{Same layout as Figure~\ref{fig:curriculum_vs_nocurr_per_L}, for the full sweep $L \in \{96, 104, 112, 120, 128, 256\}$.}
\label{fig:curriculum_vs_nocurr_per_L_full}
\end{figure}

Figure~\ref{fig:curriculum_vs_nocurr_per_L_full} shows training-compute-vs-accuracy curves for the complete sweep of target lookaheads $L \in \{96, 104, 112, 120, 128, 256\}$ from Section~\ref{sec:results:curriculum_necessary}. The body of the paper shows three representative panels ($L{=}96, 128, 256$) for compactness; the additional intermediate values ($L{=}104, 112, 120$) included here exhibit the same qualitative pattern: training with curriculum reaches the $98\%$ threshold $1.25$--$2.27\times$ faster than training without curriculum at every $L \le 128$ (Table~\ref{tab:curr_vs_direct}).

\subsection{Joint Curriculum vs.\ Fixed Attribute Vocabulary (Section~\ref{sec:results:vocab_ablation})}
\label{app:vocab_ablation_table}

Table~\ref{tab:vocab_ablation_compute} reports, for the joint curriculum and for the lookahead-only curriculum with a fixed attribute vocabulary (Section~\ref{sec:results:vocab_ablation}), the cumulative compute at which each configuration completes stage $L$, and the stage completed at the shared budget.

\begin{table}[t]
\centering
\small
\begin{tabular}{crr}
\toprule
Stage $L$ & Joint curriculum & Lookahead only \\
\midrule
8 & 144 & 401 \\
16 & 983 & 2.54K \\
24 & 3.14K & 4.30K \\
32 & 6.19K & 7.66K \\
48 & 13.6K & 20.6K \\
64 & 23.5K & 43.7K \\
80 & 37.7K & 61.6K \\
96 & 67.3K & 88.8K \\
112 & 74.3K & -- \\
128 & 87.5K & -- \\
\midrule
$L$ at 100K & \textbf{128} & 96 \\
\bottomrule
\end{tabular}
\caption{Cumulative training compute (PFLOPs) at which each configuration completes stage $L$ for \texttt{Qwen3-0.6B} with $s{=}8$, and the stage completed within a shared budget of 100K PFLOPs. The joint curriculum grows the attribute vocabulary as $|\mathcal{V}_k| = 2(L_k + 1)$, and the lookahead-only curriculum holds it fixed at its maximum size. A dash (--) means the model never reached that lookahead during training. The fixed-vocabulary configuration ends with $L{=}96$ completed after 107K PFLOPs. PFLOPs are computed as $6N$ per token with actual parameter counts.}
\label{tab:vocab_ablation_compute}
\end{table}

\subsection{Model Size (Section~\ref{sec:results:model_size})}
\label{app:model_size_table}

Table~\ref{tab:model_size_final} reports, for the four \texttt{Pythia} sizes trained under the curriculum with $s{=}1$, the achieved lookahead $L$ at the end of training and the cumulative training compute at which that stage is completed. Table~\ref{tab:model_size_results} reports the stage completed at fixed compute budgets. Both use the convention of Section~\ref{sec:experiments} (the reported $L$ is the last stage completed).

\begin{table}[H]
\centering
\small
\begin{tabular}{lcr}
\toprule
Model & Achieved lookahead $L$ & Completed at (PFLOPs) \\
\midrule
\texttt{Pythia-160M} & 9 & 921K \\
\texttt{Pythia-410M} & 46 & 1.72M \\
\texttt{Pythia-1.4B} & 75 & 4.97M \\
\texttt{Pythia-2.8B} & 133 & 2.55M \\
\bottomrule
\end{tabular}
\caption{Achieved lookahead $L$ at the end of training for each \texttt{Pythia} size under the curriculum with $s{=}1$, and the cumulative training compute (PFLOPs, computed as $6N$ per token) at which that stage is completed.}
\label{tab:model_size_final}
\end{table}

\begin{table}[H]
\centering
\small
\setlength{\tabcolsep}{4pt}
\begin{tabular}{lcccccccrr}
\toprule
& \multicolumn{6}{c}{$L$ completed at cumulative PFLOPs} & \multicolumn{3}{c}{End of training} \\
\cmidrule(lr){2-7} \cmidrule(lr){8-10}
Model & 100K & 200K & 300K & 500K & 750K & 1M & $L$ & PFLOPs & Tokens \\
\midrule
\texttt{Pythia-160M} & 7 & 7 & 8 & 8 & 8 & 9 & 9 & 995,662 & 1,022B \\
\texttt{Pythia-410M} & 30 & 34 & 37 & 40 & 42 & 44 & 46 & 2,017,056 & 828B \\
\texttt{Pythia-1.4B} & 34 & 41 & 46 & 52 & 56 & 58 & 75 & 5,000,829 & 589B \\
\texttt{Pythia-2.8B} & 31 & 46 & 60 & 73 & 84 & 94 & 133 & 2,561,267 & 154B \\
\bottomrule
\end{tabular}
\caption{Curriculum stage completed by each \texttt{Pythia} size ($s{=}1$) at fixed cumulative training compute and at the end of training. \texttt{Pythia-410M} leads \texttt{Pythia-1.4B} only below ${\sim}30$K PFLOPs, and \texttt{Pythia-2.8B} overtakes \texttt{Pythia-1.4B} between 100K and 200K PFLOPs. \texttt{Pythia-2.8B} completes $L{=}130$ at 2.40M PFLOPs and $L{=}133$ at 2.55M PFLOPs; \texttt{Pythia-1.4B} completes $L{=}75$ at 4.97M PFLOPs.}
\label{tab:model_size_results}
\end{table}

Table~\ref{tab:model_size_compute} reports the cumulative compute at which each \texttt{Pythia} size completes stage $L$.

\begin{table}[H]
\centering
\small
\setlength{\tabcolsep}{4pt}
\begin{tabular}{lrrrrrrrr}
\toprule
Model & 8 & 16 & 24 & 32 & 48 & 64 & 96 & 128 \\
\midrule
\texttt{Pythia-160M} & 253K & -- & -- & -- & -- & -- & -- & -- \\
\texttt{Pythia-410M} & 748 & 7.64K & 34.3K & 134K & -- & -- & -- & -- \\
\texttt{Pythia-1.4B} & 1.07K & 7.75K & 28.2K & 73.6K & 385K & 1.65M & -- & -- \\
\texttt{Pythia-2.8B} & 2.62K & 15.2K & 33.6K & 102K & 212K & 347K & 1.04M & 2.30M \\
\bottomrule
\end{tabular}
\caption{Cumulative training compute (PFLOPs) at which each \texttt{Pythia} model completes stage $L$ under the curriculum with $s{=}1$. A dash (--) means the model never reached that lookahead during training. Training ends with $L{=}9$ (160M, 996K PFLOPs), $46$ (410M, 2.02M), $75$ (1.4B, 5.00M) and $133$ (2.8B, 2.56M) completed. PFLOPs are computed as $6N$ per token with actual parameter counts.}
\label{tab:model_size_compute}
\end{table}

Table~\ref{tab:model_size_tokens} reports the same stage completions in training tokens.

\begin{table}[H]
\centering
\small
\setlength{\tabcolsep}{4pt}
\begin{tabular}{lrrrrrrrr}
\toprule
Model & 8 & 16 & 24 & 32 & 48 & 64 & 96 & 128 \\
\midrule
\texttt{Pythia-160M} & 260 & -- & -- & -- & -- & -- & -- & -- \\
\texttt{Pythia-410M} & 0.31 & 3.1 & 14.1 & 55.1 & -- & -- & -- & -- \\
\texttt{Pythia-1.4B} & 0.13 & 0.91 & 3.3 & 8.7 & 45.4 & 194.8 & -- & -- \\
\texttt{Pythia-2.8B} & 0.16 & 0.91 & 2.0 & 6.2 & 12.7 & 20.9 & 62.5 & 138.4 \\
\bottomrule
\end{tabular}
\caption{Training tokens (billions) at which each \texttt{Pythia} model completes stage $L$ under the curriculum with $s{=}1$; the same stage completions as Table~\ref{tab:model_size_compute}, in tokens rather than PFLOPs. A dash (--) means the model never reached that lookahead during training.}
\label{tab:model_size_tokens}
\end{table}

\subsection{Initialization (Section~\ref{sec:results:initialization})}
\label{app:init_table}

Table~\ref{tab:init_compute} reports, for \texttt{Pythia-1.4B} and \texttt{Qwen3-0.6B} initialized from the released checkpoint or from random weights (Section~\ref{sec:results:initialization}), the cumulative compute at which each condition completes stage $L$ and the stage completed at two budgets.

\begin{table}[t]
\centering
\small
\begin{tabular}{rrrrr}
\toprule
 & \multicolumn{2}{c}{\texttt{Pythia-1.4B}} & \multicolumn{2}{c}{\texttt{Qwen3-0.6B}} \\
\cmidrule(lr){2-3}\cmidrule(lr){4-5}
Stage $L$ & pretrained & random & pretrained & random \\
\midrule
8   & 1.07K & 79.7K & 618   & 74.8K \\
16  & 7.75K & 636K  & 2.35K & 285K \\
24  & 28.2K & 1.64M & 5.78K & 618K \\
32  & 73.6K & 4.41M & 10.8K & 1.36M \\
48  & 385K  & --    & 25.4K & 2.22M \\
64  & 1.65M & --    & 48.0K & -- \\
96  & --    & --    & 121K  & -- \\
128 & --    & --    & 228K  & -- \\
192 & --    & --    & 628K  & -- \\
244 & --    & --    & 1.49M & -- \\
\midrule
$L$ at 300K & 46 & 13 & 145 & 16 \\
$L$ at 1.5M & \textbf{63} & 23 & \textbf{244} & 35 \\
\bottomrule
\end{tabular}
\caption{Cumulative training compute (PFLOPs) at which each condition completes stage $L$ under the curriculum with $s{=}1$, and the stage completed at 300K and 1.5M PFLOPs. A dash (--) means the model never reached that lookahead during training. Training from random initialization ends with $L{=}35$ completed (\texttt{Pythia-1.4B}, 6.02M PFLOPs) and $L{=}61$ completed (\texttt{Qwen3-0.6B}, 4.82M PFLOPs). PFLOPs are computed as $6N$ per token with actual parameter counts.}
\label{tab:init_compute}
\end{table}

\begin{figure}[H]
\centering
\includegraphics[width=0.8\linewidth]{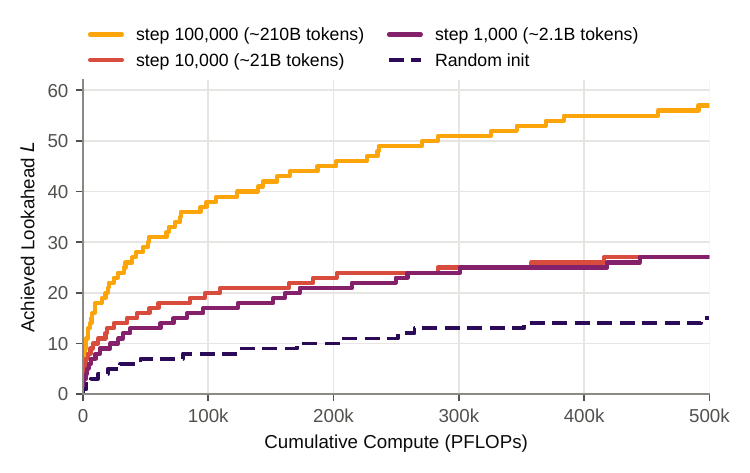}
\caption{Achieved lookahead $L$ vs.\ cumulative training compute (PFLOPs) for \texttt{Pythia-1.4B} fine-tuned from intermediate pretraining checkpoints, each stopped at 500K~PFLOPs. The dashed curve is the random-initialization trajectory from Section~\ref{sec:results:initialization}.}
\label{fig:pythia_revisions}
\end{figure}

\paragraph{Intermediate pretraining checkpoints.}
We test whether the benefit grows with the amount of pretraining within a single model. We fine-tune \texttt{Pythia-1.4B} from three intermediate checkpoints released by \citet{biderman2023pythia}, at pretraining steps 1,000, 10,000 and 100,000 (about 2.1B, 21B and 210B tokens).
Each trajectory follows the curriculum with $s{=}1$, using the pretrained \texttt{Pythia-1.4B} batch size and learning rate (192 and $2.7\times10^{-5}$) with no warmup, and stops at 500K~PFLOPs.
Later checkpoints advance faster (Figure~\ref{fig:pythia_revisions}, Table~\ref{tab:pythia_revisions}). The step-100,000 checkpoint completes $L{=}24$ with about $9\times$ less compute than the step-1,000 checkpoint and reaches $L{=}57$ within the budget, while both earlier checkpoints stop at $L{=}27$.
The advantage of step 10,000 over step 1,000 shrinks as $L$ grows, while the advantage of step 100,000 over step 10,000 widens.
Even the step-1,000 checkpoint completes $L{=}8$ with about $8\times$ less compute than random initialization.

\begin{table}[H]
\centering
\small
\begin{tabular}{rrrr}
\toprule
Stage $L$ & Step 1,000 & Step 10,000 & Step 100,000 \\
 & (${\sim}$2.1B tokens) & (${\sim}$21B) & (${\sim}$210B) \\
\midrule
8  & 9.95K & 3.79K & 1.04K \\
16 & 83.0K & 43.1K & 7.25K \\
24 & 259K  & 203K  & 27.7K \\
32 & --    & --    & 66.7K \\
48 & --    & --    & 235K \\
\midrule
$L$ at 100K & 17 & 20 & 38 \\
$L$ at 300K & 24 & 25 & 51 \\
$L$ at 500K & 27 & 27 & 57 \\
\bottomrule
\end{tabular}
\caption{Cumulative training compute (PFLOPs) at which \texttt{Pythia-1.4B}, fine-tuned from intermediate pretraining checkpoints, completes stage $L$ under the curriculum with $s{=}1$, and the stage completed at three budgets. Column headers give the pretraining step and the approximate number of pretraining tokens. A dash (--) means the model never reached that lookahead during training. Each trajectory ends at 500K~PFLOPs.}
\label{tab:pythia_revisions}
\end{table}

\subsection{LoRA vs.\ Full Fine-Tuning}
\label{app:lora}

We compare full fine-tuning, used in all other experiments, with LoRA \citep{hu2022lora} at ranks $r{=}8$ and $r{=}64$. All three conditions train \texttt{Qwen3-0.6B} under the curriculum with $s{=}8$ up to $L{=}128$, using the step-size sweep settings of Appendix~\ref{app:hyperparams} without warmup.
Adapters are applied to all attention and MLP projections, with $\alpha{=}2r$ and dropout 0.1.
Because the frozen base weights need no weight gradients, we count LoRA compute as $4N + 6N_{\text{LoRA}}$ per token, where $N_{\text{LoRA}}$ is the number of adapter parameters.
Both ranks reach $L{=}128$ with roughly the same compute as full fine-tuning ($0.95\times$ for $r{=}8$ and $0.85\times$ for $r{=}64$) but need more optimizer steps ($1.56\times$ and $1.13\times$; Table~\ref{tab:lora}).
Each condition uses a single seed, so these ratios are approximate.
We use full fine-tuning elsewhere because it reaches the same lookahead with similar compute and does not require choosing a LoRA rank.

\begin{table}[H]
\centering
\small
\begin{tabular}{rrrrrrr}
\toprule
 & \multicolumn{3}{c}{Compute (PFLOPs)} & \multicolumn{3}{c}{Optimizer steps} \\
\cmidrule(lr){2-4}\cmidrule(lr){5-7}
Stage $L$ & Full FT & $r{=}8$ & $r{=}64$ & Full FT & $r{=}8$ & $r{=}64$ \\
\midrule
32  & 4.25K & 7.97K & 4.49K & 3,250  & 8,575  & 4,375 \\
64  & 25.6K & 34.3K & 22.1K & 10,700 & 22,175 & 12,600 \\
96  & 66.5K & 60.4K & 45.2K & 19,550 & 31,000 & 19,475 \\
128 & 116K  & 110K  & 98.1K & 27,575 & 43,125 & 31,200 \\
\bottomrule
\end{tabular}
\caption{Cumulative training compute and optimizer steps at which \texttt{Qwen3-0.6B} completes stage $L$ under the curriculum with $s{=}8$, with full fine-tuning (Full FT) and with LoRA at rank $r$. LoRA compute is counted as $4N + 6N_{\text{LoRA}}$ per token.}
\label{tab:lora}
\end{table}

\FloatBarrier

\section{Functional forms for achieved lookahead}
\label{app:curve-fit}

\begin{figure}[t]
    \centering
    \includegraphics[width=0.8\linewidth]{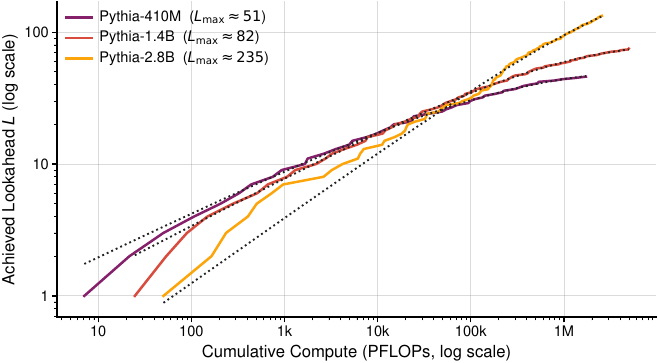}
    \caption{Saturating exponential fits (black dotted, over the observed range) to the achieved-lookahead trajectories of \texttt{Pythia-410M}, \texttt{-1.4B}, and \texttt{-2.8B} trained with $s{=}1$ (log--log). Legends give each fitted ceiling $L_{\max}$ (Table~\ref{tab:fit-eqns}).}
    \label{fig:lmax_scaling}
\end{figure}

We fit achieved lookahead $L$ as a function of cumulative training compute $C$ for the two trajectories of Section~\ref{sec:results:scaling}, \texttt{Qwen3-0.6B} and \texttt{Pythia-1.4B}, and for the less trained \texttt{Pythia-410M} and \texttt{Pythia-2.8B}, using the power law and saturating exponential defined there. This appendix reports the full comparison (RMSE with confidence intervals, AICc, and held-out prediction of both cumulative lookahead and per-stage training iterations), along with the per-size saturating exponential fits underlying the ceiling analysis of Section~\ref{sec:results:scaling} (Figure~\ref{fig:lmax_scaling}).

\paragraph{Data and procedure.} Stage completions (one point per stage, $L \in [1, L_{\text{end}}]$, at the compute where the $98\%$ gate is passed) yield $n = 244$ points for \texttt{Qwen3-0.6B} and $n = 46$, $75$ and $133$ for \texttt{Pythia-410M}, \texttt{-1.4B} and \texttt{-2.8B}. We fit each form with SciPy \texttt{differential\_evolution} (Sobol init, popsize=30, maxiter=200, seed=42) polished by Levenberg--Marquardt, retaining the lowest-RSS solution. Confidence intervals on RMSE are 95\% moving-block bootstrap intervals (block length ${\approx}\,n^{1/3}$, 300--500 resamples), which partly account for the serial correlation of a single trajectory; with residual autocorrelation persisting over many lags they should be read as descriptive, not as calibrated 95\% intervals. $\Delta$AICc treats the stage points as independent, which they are not (residual autocorrelation exceeds $0.8$ at lag 1), so we read it as a ranking of in-sample fit rather than as a calibrated test; the formal comparison is the held-out test on per-stage training iterations in Appendix~\ref{app:increment-test}. Because the power-law offset is at its bound $L_0 = 0$ in every fit and the saturating exponential approaches $aC^{k}$ as $C_0$ grows, the in-sample comparison is one-sided by construction and only the size of the RMSE gap is informative. Held-out comparisons fit both forms on the first $70\%$ of the stage points, ordered by compute, and evaluate their predictions on the remainder (Table~\ref{tab:curve_heldout_all}).

\begin{table}[H]
\centering
\small
\begin{tabular}{lrrrr}
\toprule
& \multicolumn{2}{c}{Held-out RMSE} & & \\
\cmidrule(lr){2-3}
Trajectory & Power law & Sat.\ exp. & Ratio & $p$ \\
\midrule
\texttt{Qwen3-0.6B} & 23.88 & 17.08 & 1.40 & $7{\times}10^{-10}$ \\
\texttt{Pythia-410M} & 8.23 & 0.84 & 9.80 & $0.018$ \\
\texttt{Pythia-1.4B} & 16.14 & 0.68 & 23.77 & $1.5{\times}10^{-4}$ \\
\texttt{Pythia-2.8B} & 9.25 & 3.72 & 2.49 & $0.65$ \\
\bottomrule
\end{tabular}
\caption{Held-out prediction error for achieved lookahead $L$ on all four $s{=}1$ trajectories, fitting on the first $70\%$ of stage completions and predicting the remainder. Ratio is power-law over saturating-exponential RMSE. $p$: significance of the saturating exponential's advantage under the per-stage iteration test of Appendix~\ref{app:increment-test}, two-sided.}
\label{tab:curve_heldout_all}
\end{table}

\begin{table}[H]
\centering
\small
\setlength{\tabcolsep}{4pt}
\resizebox{\textwidth}{!}{%
\begin{tabular}{lccccc}
\toprule
Trajectory & $n$ & $L_{\text{end}}$ & Power-law RMSE [95\% CI] & Saturating exponential RMSE [95\% CI] & $\Delta$AICc \\
\midrule
\texttt{Qwen3-0.6B}  & 244 & 244 & $6.10$ $[3.27, 6.75]$ & $2.22$ $[1.57, 2.35]$ & 493 \\
\texttt{Pythia-410M} & 46  & 46  & $1.93$ $[0.71, 2.05]$ & $0.47$ $[0.32, 0.53]$ & 130 \\
\texttt{Pythia-1.4B} & 75  & 75  & $2.98$ $[1.69, 3.24]$ & $0.53$ $[0.40, 0.57]$ & 259 \\
\texttt{Pythia-2.8B} & 133 & 133 & $2.67$ $[1.92, 3.16]$ & $2.15$ $[1.46, 2.54]$ & 57 \\
\bottomrule
\end{tabular}}
\caption{Fit comparison for the four long $s{=}1$ trajectories. $\Delta$AICc is (power-law $-$ saturating exponential); positive favors the saturating exponential, and since both forms have three parameters it reflects fit quality alone. RMSE confidence intervals are 95\% moving-block bootstrap. Both in-sample criteria favor the saturating exponential on every trajectory. Held-out prediction is compared in Table~\ref{tab:curve_heldout_all} and Appendix~\ref{app:increment-test}.}
\label{tab:fit-gof}
\end{table}

\begin{table}[H]
\centering
\small
\begin{tabular}{lccc}
\toprule
Trajectory & $L_{\max}$ [95\% CI] & $C_0$ (PFLOPs) & $k$ \\
\midrule
\texttt{Qwen3-0.6B}  & $323.6$ $[304, 404]$ & $7.7{\times}10^{5}$ & 0.547 \\
\texttt{Pythia-410M} & $51.2$ $[49.2, 56.9]$ & $1.4{\times}10^{5}$ & 0.341 \\
\texttt{Pythia-1.4B} & $81.9$ $[77.8, 83.0]$ & $5.0{\times}10^{5}$ & 0.371 \\
\texttt{Pythia-2.8B} & $235.1$ $[175, 315]$ & $3.8{\times}10^{6}$ & 0.497 \\
\bottomrule
\end{tabular}
\caption{Fitted saturating exponential parameters, $L = L_{\max}\bigl(1 - \exp(-(C/C_0)^{k})\bigr)$, for the four long $s{=}1$ trajectories. $L_{\max}$ is the fitted ceiling plotted in Figure~\ref{fig:lmax_scaling}. Brackets give 95\% moving-block bootstrap percentile intervals for $L_{\max}$ (block length ${\approx}\,n^{1/3}$, 300 resamples). Their lower ends are stable under longer blocks, while the upper ends for \texttt{Qwen3-0.6B} and \texttt{Pythia-2.8B} widen substantially, so those upper bounds are indicative only.}
\label{tab:fit-eqns}
\end{table}

\paragraph{What the comparison shows.} On every trajectory the saturating exponential attains lower RMSE, is preferred by AICc ($\Delta\text{AICc} = 57$--$493$; Table~\ref{tab:fit-gof}) and predicts the held-out stages with lower RMSE (Table~\ref{tab:curve_heldout_all}). The in-sample gap is largest on the three trajectories nearest their fitted ceilings and small on \texttt{Pythia-2.8B} (RMSE $2.67$ vs.\ $2.15$), which sits at only ${\sim}57\%$ of its fitted ceiling and whose RMSE intervals overlap; the two forms separate more cleanly as a model nears its ceiling. Appendix~\ref{app:increment-test} reports a significance test on per-stage training iterations. Figure~\ref{fig:lmax_scaling} plots the per-size saturating exponential fits (parameters in Table~\ref{tab:fit-eqns}); the resulting ceilings ($L_{\max} \approx 51$, $82$, $235$) underlie the ceiling analysis in Section~\ref{sec:results:scaling}. The bootstrap intervals in Table~\ref{tab:fit-eqns} show that the lower bound of each ceiling is well determined, while the upper bound is poorly determined for the two trajectories furthest from saturation, \texttt{Qwen3-0.6B} and \texttt{Pythia-2.8B}.
\texttt{Pythia-160M} plateaus at $L{=}9$ (Section~\ref{sec:results:model_size}) and is excluded, since neither form is meaningful for a capacity-limited trajectory.

\subsection[Held-out prediction of per-stage training iterations]{Held-out prediction of per-stage training iterations}
\label{app:increment-test}

A Diebold--Mariano test on the held-out residuals of $L(C)$ is not valid for these trajectories. Its variance estimate assumes a weakly dependent loss differential, whereas the held-out residuals along a single trajectory share one extrapolation error and are strongly autocorrelated (lag-1 autocorrelation $0.95$ for \texttt{Qwen3-0.6B} and $0.92$ for \texttt{Pythia-2.8B}; Table~\ref{tab:increment_diag}). We therefore compare the two forms on a quantity whose residuals are approximately independent, the number of training iterations required to complete each stage, $\Delta I(L) = I(L) - I(L-1)$ with $I(0) = 0$, where $I(L)$ is the cumulative number of optimizer steps at the completion of stage $L$. We use iterations rather than compute because the compute cost of a stage is its iteration count multiplied by the tokens per step, and tokens per step grow with $L$ as the sequence length grows, so per-stage iterations isolate the learning dynamics from that growth. Each form implies a per-stage iteration count $g(L) - g(L-1)$, where $g$ is the inverse of the form fitted with cumulative iterations as the independent variable. We fit both forms to $\log \Delta I(L)$ by least squares on the first $70\%$ of stages, ordered by compute, and predict the remaining stages. Table~\ref{tab:increment_diag} shows that on per-stage iterations the residuals of the saturating exponential are consistent with independence on three trajectories and marginal on \texttt{Pythia-1.4B} ($p = 0.046$).

\begin{table}[H]
\centering
\small
\begin{tabular}{lcccc}
\toprule
& \multicolumn{2}{c}{Cumulative lookahead $L(C)$} & \multicolumn{2}{c}{Per-stage iterations $\Delta I(L)$} \\
\cmidrule(lr){2-3} \cmidrule(lr){4-5}
Trajectory & lag-1 autocorr. & Ljung--Box $p$ & lag-1 autocorr. & Ljung--Box $p$ \\
\midrule
\texttt{Qwen3-0.6B} & $0.95$ & $2{\times}10^{-199}$ & $-0.14$ & $0.15$ \\
\texttt{Pythia-410M} & $0.25$ & $0.36$ & $-0.28$ & $0.36$ \\
\texttt{Pythia-1.4B} & $0.33$ & $0.08$ & $-0.33$ & $0.046$ \\
\texttt{Pythia-2.8B} & $0.92$ & $4{\times}10^{-94}$ & $-0.10$ & $0.13$ \\
\bottomrule
\end{tabular}
\caption{Residual dependence of the saturating exponential fit on cumulative lookahead and on per-stage training iterations (log units). Ljung--Box tests use five lags. Per-stage iterations remove the serial dependence that invalidates a Diebold--Mariano test on cumulative lookahead.}
\label{tab:increment_diag}
\end{table}

For each held-out stage we form the loss differential $d_L = e_{\mathrm{PL}}(L)^2 - e_{\mathrm{SE}}(L)^2$ in log units, which is positive when the saturating exponential predicts that stage better. Because the advantage may grow with depth, we regress $d_L$ on the centered stage index with Newey--West standard errors (Bartlett kernel, bandwidth $\lfloor 4(T/100)^{2/9} \rfloor$) and report the intercept $\alpha$, which equals the mean of $d_L$, and the slope $\beta$ (Table~\ref{tab:increment_heldout}). All $p$-values are two-sided and are stable for bandwidths from $0$ to four times this rule. The saturating exponential has lower held-out RMSE on per-stage iterations for three trajectories and ties on \texttt{Pythia-2.8B}. The advantage is significant for \texttt{Qwen3-0.6B}, \texttt{Pythia-410M} and \texttt{Pythia-1.4B}, and on all three it grows with depth ($\beta > 0$). Across train/held-out splits from $50\%$ to $80\%$ of the stages, $\alpha$ is significant at every split for \texttt{Qwen3-0.6B} and \texttt{Pythia-410M}, at the $70\%$ split only for \texttt{Pythia-1.4B}, and at no split for \texttt{Pythia-2.8B}. The \texttt{Pythia-1.4B} result is sensitive to the split because that trajectory's per-stage costs vary by two orders of magnitude between adjacent stages, so a held-out set of $8$ to $24$ stages is dominated by a few of them. For \texttt{Pythia-2.8B} the two forms predict per-stage iterations identically, because the saturating exponential fitted to that trajectory's iterations has no identified ceiling and reduces to the power law. Per-stage iterations are a noisy quantity, with held-out RMSE near $1$ in log units for both forms, so this test has less power than the comparison on cumulative lookahead but rests on assumptions the data satisfy.

\begin{table}[H]
\centering
\small
\setlength{\tabcolsep}{4pt}
\resizebox{\textwidth}{!}{%
\begin{tabular}{lcccccccc}
\toprule
& & \multicolumn{3}{c}{Held-out RMSE ($\log \Delta I$)} & \multicolumn{2}{c}{Mean advantage $\alpha$} & \multicolumn{2}{c}{Trend $\beta$ (per stage)} \\
\cmidrule(lr){3-5} \cmidrule(lr){6-7} \cmidrule(lr){8-9}
Trajectory & $T$ & Sat.\ exp. & Power law & Ratio & Estimate [95\% CI] & $p$ & Estimate [95\% CI] & $p$ \\
\midrule
\texttt{Qwen3-0.6B} & 73 & $0.93$ & $0.96$ & $1.04$ & $0.068$ $[0.049, 0.087]$ & $7{\times}10^{-10}$ & $0.0025$ $[0.0016, 0.0034]$ & $1{\times}10^{-6}$ \\
\texttt{Pythia-410M} & 14 & $1.14$ & $1.90$ & $1.67$ & $2.33$ $[0.47, 4.18]$ & $0.018$ & $0.51$ $[0.08, 0.95]$ & $0.025$ \\
\texttt{Pythia-1.4B} & 23 & $1.06$ & $1.76$ & $1.66$ & $1.97$ $[1.08, 2.86]$ & $1.5{\times}10^{-4}$ & $0.18$ $[0.05, 0.32]$ & $0.010$ \\
\texttt{Pythia-2.8B} & 40 & $1.13$ & $1.13$ & $1.00$ & $6{\times}10^{-6}$ $[-2{\times}10^{-5}, 3{\times}10^{-5}]$ & $0.65$ & $-9{\times}10^{-7}$ $[-3{\times}10^{-6}, 1{\times}10^{-6}]$ & $0.37$ \\
\bottomrule
\end{tabular}}
\caption{Held-out prediction of per-stage training iterations. $T$ is the number of held-out stages. $\alpha$ and $\beta$ are the intercept and slope of $d_L$ regressed on the centered stage index, with Newey--West standard errors; $\alpha > 0$ favors the saturating exponential and $\beta > 0$ indicates an advantage that grows with depth.}
\label{tab:increment_heldout}
\end{table}

\begin{figure}[H]
    \centering
    \includegraphics[width=\linewidth]{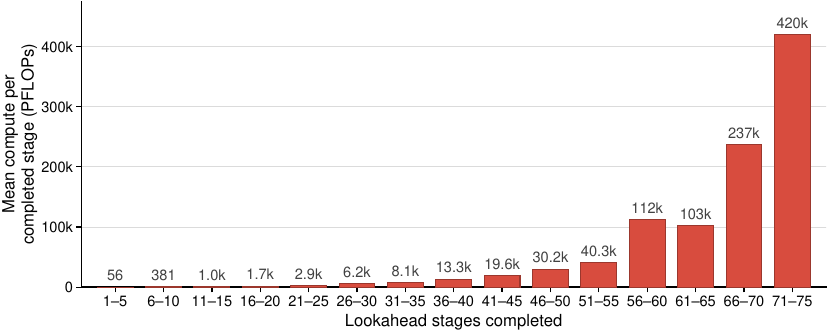}
    \caption[Compute per curriculum stage for \texttt{Pythia-1.4B}.]{Mean training compute per completed curriculum stage for \texttt{Pythia-1.4B} with $s{=}1$. Each bar averages five consecutive completed stages; labels show the mean in PFLOPs. Compute for a stage is measured between consecutive stage completions, with the first stage measured from the start of training. The unfinished stage $L{=}76$ is excluded.}
    \label{fig:per_stage_cost_pythia14b}
\end{figure}

\begin{samepage}
\subsection[Compute per curriculum stage]{Compute per curriculum stage}
\label{app:per_stage_compute}

Figure~\ref{fig:per_stage_cost_pythia14b} shows the mean compute per completed curriculum stage for \texttt{Pythia-1.4B} with $s=1$, averaged over consecutive groups of five stages. The last five completed stages, $L{=}71$--$75$, consume $2.10$M PFLOPs, approximately $42\%$ of its 5M-PFLOP budget. Thus, nearly half the training compute produces only five additional stages of progress, illustrating the sharply diminishing returns to further training in this regime. Within this budget, the trajectory reaches $L{=}75$, $91\%$ of the fitted asymptote $L_{\max}\approx82$, without having plateaued. The flattening captured by the saturating exponential is therefore accompanied by a substantial observed increase in the compute required for further curriculum progress.

\end{samepage}

\section{Downstream Evaluation Methodology}
\label{app:eval_methodology}

\paragraph{Next-step proof prediction.} On the synthetic next-step prediction task (Section~\ref{sec:methods:task}) we use greedy autoregressive decoding, and the first whitespace- or punctuation-delimited word of the generated continuation is compared case-insensitively with any accepted gold attribute.

We evaluate on 14 reasoning benchmarks (full list in Table~\ref{tab:downstream_curriculum_small}). Thirteen are scored by log-likelihood over a fixed candidate label set; the open-ended \texttt{NLGraph} benchmark uses its authors' per-task scorers. Samples are drawn with a fixed random seed, and the number of items per benchmark is listed in Table~\ref{tab:benchmarks}.

\paragraph{Why constrained log-likelihood rather than CoT.} Several of these benchmarks are designed for chain-of-thought or free-form decoding. Our curriculum-trained models do not produce verbose CoT natively. The curriculum task contains no signal for natural-language reasoning style, so unconstrained free-form decoding would entangle output style with underlying capability. For benchmarks with a fixed answer set we therefore use log-likelihood scoring over the candidate labels, isolating reasoning from format-following. This is the same convention used by \texttt{lm-evaluation-harness}~\citep{eval-harness} for its \texttt{loglikelihood} request type.

\paragraph{Fused single-pass scoring.} For each example we run the model once on the prompt and compute the per-token log-probabilities of every candidate label. When a label tokenizes to a single token (the common case for True/False/Unknown, multiple-choice letters, etc.) we read its log-prob directly from the next-token distribution. For multi-token labels we fall back to teacher-forced log-likelihood with the same approach as \texttt{lm-evaluation-harness}: walk forward to the first divergent token between $\mathrm{tokenize}(\text{prefix})$ and $\mathrm{tokenize}(\text{prefix}+\text{label})$, then sum log-probs of the suffix tokens. This handles BPE boundary merges correctly and remains a single forward pass per choice.

\paragraph{Reported metric.} The headline accuracy reported in Table~\ref{tab:downstream_curriculum_small} is log-likelihood accuracy: argmax over candidate labels by total log-probability. This matches the dominant convention for log-likelihood-scored benchmarks in the literature.

\paragraph{Per-benchmark prompting protocol.} For each benchmark we adapt its canonical LLM evaluation protocol to a constrained-log-likelihood setting: we keep the canonical demonstration count and prompt format where compatible, and substitute Direct prompting where the canonical convention uses chain-of-thought. The same protocol is applied to the base model, every curriculum checkpoint, the no-curriculum baseline, and the instruction-only control for each benchmark. Per-benchmark choices are summarized in Table~\ref{tab:eval_protocol}.

\begin{table}[H]
\centering\small
\begin{tabular}{lll}
\toprule
Benchmark & Canonical LLM protocol & Our protocol \\
\midrule
\multicolumn{3}{l}{\emph{Reasoning benchmarks}} \\
ProofWriter OWA & none & 0-shot Direct \\
ProofWriter CWA & none & 0-shot Direct \\
ParaRules & none & 0-shot Direct \\
ZebraLogic-MC & none & 3-shot Direct \\
PrOntoQA & 8-shot CoT~\citep{saparov2023testing} & 8-shot Direct \\
FOLIO & 8-shot~\citep{han2024folio} & 8-shot Direct \\
JustLogic & 3-shot~\citep{chen2025justlogiccomprehensivebenchmarkevaluating} & 3-shot Direct \\
ProverQA & 2-shot~\citep{ICLR2025_862819c2} & 2-shot Direct \\
CLUTRR & none & 5-shot Direct \\
LogicBench-BQA & 0-shot CoT~\citep{parmar2024logicbench} & 3-shot Direct \\
RuleTaker & none & 5-shot Direct \\
Web of Lies & 3-shot~\citep{suzgun-etal-2023-challenging} & 3-shot Direct \\
LogiQA & 0-shot~\citep{zhong2023agieval} & 0-shot Direct \\
NLGraph & 1--5-shot~\citep{wang2023nlgraph} & 1--5-shot, generation \\
\midrule
\multicolumn{3}{l}{\emph{General-capability benchmarks}} \\
MMLU & 0-shot~\citep{eval-harness} & 0-shot \\
IFEval & 0-shot~\citep{zhou2023instruction} & 0-shot, generation \\
ARC-E & none & 0-shot \\
ARC-C & none & 0-shot \\
HellaSwag & none & 0-shot \\
WinoGrande & none & 0-shot \\
PIQA & none & 0-shot \\
BoolQ & none & 0-shot \\
LAMBADA & none & 0-shot \\
\bottomrule
\end{tabular}
\caption{Per-benchmark prompting protocol. Where a benchmark has a canonical zero-shot protocol, we keep it; where it has a canonical few-shot protocol, we use its shot count; where the canonical protocol uses chain-of-thought or no LLM protocol exists, we use Direct prompting. All demonstrations show the question and label only and are removed from the scored items. The seven general-capability benchmarks without a canonical protocol use the \texttt{lm-evaluation-harness} default of zero-shot.}
\label{tab:eval_protocol}
\end{table}

\paragraph{Sampling and stratification.} Item counts per benchmark are listed in Table~\ref{tab:benchmarks}. For benchmarks with subsets (per-depth, per-task, or per-puzzle-size), we draw a fixed-seed random sample within each subset rather than a deterministic prefix. \texttt{ProofWriter OWA} and \texttt{ParaRules} are label-balanced within each depth; \texttt{PrOntoQA} and \texttt{Web of Lies} use 500 items per depth; \texttt{NLGraph} uses 125 items per sub-task.
\paragraph{NLGraph (generation).} \texttt{NLGraph} contains 8 sub-tasks (shortest path, Hamilton, connectivity, etc.) and the canonical scoring is per-task: path validation for Hamilton, weight match for shortest path, regex-based answer extraction for others. We use the authors' canonical \texttt{k-shot-prompt.txt} files (with task-dependent $k$, no CoT) and their per-task scorers ported from the official repo. The accuracy we report on \texttt{NLGraph} is the average of these per-task scores.

\section{Confidence Intervals on Downstream Results}
\label{app:downstream_ci}

Table~\ref{tab:downstream_curriculum_ci} reports the accuracies of Table~\ref{tab:downstream_curriculum_small} with 95\% confidence intervals, together with the base model and the instruction-only control, and marks every curriculum checkpoint that is significantly above training without curriculum, not only the best one per benchmark. The base model and the instruction-only control are not part of the matched comparison, and they score above curriculum $L{=}48$ on four and six of the 14 benchmarks, respectively.

\begin{table}[t]
\centering
\small
\caption{Downstream reasoning accuracy (\%) of \texttt{Qwen3-1.7B} from Table~\ref{tab:downstream_curriculum_small}, with 95\% confidence intervals, given by a simple normal approximation. Base is the non-finetuned \texttt{Qwen3-1.7B} under the same prompts, with the answer cue prefilled in its own markdown answer format. Instruction-only is trained on the instruction data alone at compute matched to curriculum $L{=}48$. $^{*}$: curriculum checkpoint significantly above no curriculum (two-sided exact McNemar test on paired items, $p<0.05$). $^\dagger$Extended by us to greater depths (Appendix~\ref{app:benchmark_descriptions}).}
\label{tab:downstream_curriculum_ci}
\resizebox{\linewidth}{!}{%
\begin{tabular}{@{}l l llll ll@{}}
\toprule
& \multirow{2}{*}[-2pt]{\textbf{Base}} & \multicolumn{4}{c}{Curriculum $L$} & \multirow{2}{*}[-2pt]{\shortstack[l]{\textbf{No}\\\textbf{curr.}}} & \multirow{2}{*}[-2pt]{\shortstack[l]{\textbf{Instr.}\\\textbf{only}}} \\
\cmidrule(lr){3-6}
\textbf{Benchmark} & & \textbf{8} & \textbf{16} & \textbf{32} & \textbf{48} & & \\
\midrule
ProofWriter OWA & 40.3{\tiny$\pm$1.8} & 41.1{\tiny$\pm$1.8} & 33.6{\tiny$\pm$1.7} & 39.3{\tiny$\pm$1.7} & 54.9{\tiny$\pm$1.8}$^{*}$ & 40.6{\tiny$\pm$1.8} & 47.8{\tiny$\pm$1.8} \\
ProofWriter CWA & 52.7{\tiny$\pm$1.8} & 66.8{\tiny$\pm$1.7}$^{*}$ & 57.2{\tiny$\pm$1.8}$^{*}$ & 67.6{\tiny$\pm$1.7}$^{*}$ & 67.9{\tiny$\pm$1.7}$^{*}$ & 55.2{\tiny$\pm$1.8} & 58.0{\tiny$\pm$1.8} \\
ParaRules & 43.8{\tiny$\pm$1.4} & 46.1{\tiny$\pm$1.4} & 36.2{\tiny$\pm$1.4} & 39.1{\tiny$\pm$1.4} & 55.1{\tiny$\pm$1.4}$^{*}$ & 50.4{\tiny$\pm$1.5} & 57.1{\tiny$\pm$1.4} \\
ZebraLogic-MC & 25.5{\tiny$\pm$2.7} & 47.4{\tiny$\pm$3.1} & 53.4{\tiny$\pm$3.1}$^{*}$ & 51.8{\tiny$\pm$3.1}$^{*}$ & 53.5{\tiny$\pm$3.1}$^{*}$ & 48.3{\tiny$\pm$3.1} & 53.2{\tiny$\pm$3.1} \\
PrOntoQA$^\dagger$ & 53.1{\tiny$\pm$1.5} & 52.5{\tiny$\pm$1.5} & 55.3{\tiny$\pm$1.5}$^{*}$ & 56.8{\tiny$\pm$1.5}$^{*}$ & 57.1{\tiny$\pm$1.5}$^{*}$ & 52.0{\tiny$\pm$1.5} & 52.2{\tiny$\pm$1.5} \\
FOLIO & 45.3{\tiny$\pm$6.8} & 45.8{\tiny$\pm$6.9}$^{*}$ & 46.3{\tiny$\pm$6.9}$^{*}$ & 45.8{\tiny$\pm$6.9}$^{*}$ & 45.8{\tiny$\pm$6.9}$^{*}$ & 41.4{\tiny$\pm$6.8} & 45.3{\tiny$\pm$6.8} \\
JustLogic & 60.9{\tiny$\pm$1.4} & 51.1{\tiny$\pm$1.4}$^{*}$ & 52.6{\tiny$\pm$1.4}$^{*}$ & 45.5{\tiny$\pm$1.4} & 53.5{\tiny$\pm$1.4}$^{*}$ & 46.1{\tiny$\pm$1.4} & 55.7{\tiny$\pm$1.4} \\
ProverQA & 48.0{\tiny$\pm$2.5} & 42.0{\tiny$\pm$2.5} & 44.4{\tiny$\pm$2.5} & 46.3{\tiny$\pm$2.5} & 47.8{\tiny$\pm$2.5} & 49.2{\tiny$\pm$2.5} & 52.0{\tiny$\pm$2.5} \\
CLUTRR & 28.6{\tiny$\pm$2.8} & 25.5{\tiny$\pm$2.7} & 27.2{\tiny$\pm$2.8} & 23.2{\tiny$\pm$2.6} & 28.6{\tiny$\pm$2.8} & 26.7{\tiny$\pm$2.7} & 22.3{\tiny$\pm$2.6} \\
LogicBench-BQA & 73.9{\tiny$\pm$2.2} & 74.2{\tiny$\pm$2.2}$^{*}$ & 72.5{\tiny$\pm$2.3} & 74.1{\tiny$\pm$2.2}$^{*}$ & 73.6{\tiny$\pm$2.2}$^{*}$ & 71.2{\tiny$\pm$2.3} & 69.5{\tiny$\pm$2.3} \\
RuleTaker & 67.8{\tiny$\pm$2.9} & 60.5{\tiny$\pm$3.0} & 60.6{\tiny$\pm$3.0} & 63.5{\tiny$\pm$3.0}$^{*}$ & 60.0{\tiny$\pm$3.0} & 58.8{\tiny$\pm$3.1} & 70.4{\tiny$\pm$2.8} \\
Web of Lies$^\dagger$ & 53.0{\tiny$\pm$1.5} & 55.5{\tiny$\pm$1.5} & 56.5{\tiny$\pm$1.5} & 57.2{\tiny$\pm$1.5} & 57.9{\tiny$\pm$1.5} & 57.8{\tiny$\pm$1.5} & 58.8{\tiny$\pm$1.5} \\
LogiQA & 39.4{\tiny$\pm$3.8} & 42.6{\tiny$\pm$3.8} & 44.9{\tiny$\pm$3.8} & 44.9{\tiny$\pm$3.8} & 42.4{\tiny$\pm$3.8} & 43.4{\tiny$\pm$3.8} & 43.2{\tiny$\pm$3.8} \\
NLGraph & 35.8{\tiny$\pm$3.0} & 41.5{\tiny$\pm$3.1} & 43.5{\tiny$\pm$3.1} & 44.3{\tiny$\pm$3.1} & 43.7{\tiny$\pm$3.1} & 46.2{\tiny$\pm$3.1} & 39.5{\tiny$\pm$3.0} \\
\bottomrule
\end{tabular}}
\end{table}

\subsection{General model capability}
\label{app:general_capability}

Table~\ref{tab:general_capability} reports the accuracies summarized in Section~\ref{sec:results:downstream}. The benchmarks are \texttt{MMLU}, \texttt{ARC-Easy} and \texttt{ARC-Challenge}, \texttt{HellaSwag}, \texttt{WinoGrande}, \texttt{PIQA}, \texttt{BoolQ}, \texttt{LAMBADA} and \texttt{IFEval}. Each is evaluated on the non-finetuned \texttt{Qwen3-1.7B}, on an instruction-only control, on the model trained without curriculum, and on four curriculum stage checkpoints. The first eight benchmarks probe knowledge and single-step commonsense rather than multi-step deductive chaining, so we do not expect the curriculum to improve them.

Because the curriculum trains the model almost exclusively on synthetic proof data, it could plausibly narrow the capabilities of the model in other respects. We evaluate \texttt{Qwen3-1.7B} on 8 standard benchmarks spanning world knowledge (\texttt{MMLU}), science QA (\texttt{ARC-Easy/Challenge}), commonsense inference (\texttt{HellaSwag}, \texttt{WinoGrande}, \texttt{PIQA}), reading comprehension (\texttt{BoolQ}) and discourse-level language modeling (\texttt{LAMBADA}), plus \texttt{IFEval} for instruction following, using an identical \texttt{lm-evaluation-harness} protocol for every model evaluated. Mean accuracy over the eight is above the non-finetuned model's at every curriculum stage (63.4 to 64.2 vs.\ 61.7; Table~\ref{tab:general_capability}). The largest single-benchmark drop below it is \texttt{WinoGrande} at $L{=}48$ (59.8 vs.\ 61.7). \texttt{IFEval} is also above the non-finetuned model at every stage (34.6 to 41.2 vs.\ 33.1). Because our training mixture also contains instruction data, a comparison against the non-finetuned model alone would leave open the possibility that this instruction data offsets damage done by the synthetic data. We therefore fine-tune the same base model on 100\% \texttt{Dolci-Instruct-SFT}, with no synthetic data and no curriculum but otherwise identical hyperparameters, and evaluate it at cumulative training compute matched to the $L{=}48$ curriculum checkpoint. It scores within one point of that checkpoint on the eight-benchmark mean (64.2 vs.\ 63.4), so training on the synthetic data adds little measurable cost on these benchmarks. The model trained without curriculum on the same data at the same compute scores essentially the same as the curriculum (63.2), so the curriculum itself adds no further cost. On \texttt{IFEval}, the $L{=}48$ checkpoint scores below the instruction-only control (41.2 vs.\ 46.4), which trains on instruction data alone, and slightly above the model trained without curriculum (40.1). Thus, we find no evidence that gains in implicit reasoning capability are at the expense of general model capability.

\begin{table}[H]
\centering
\footnotesize
\setlength{\tabcolsep}{3pt}
\caption{General model capability of \texttt{Qwen3-1.7B} across curriculum stages. Accuracy on eight standard benchmarks (zero-shot, \texttt{lm-evaluation-harness} v0.4.11, identical settings for every model) for the non-finetuned model, four curriculum stage checkpoints, the model trained without curriculum, and an instruction-only control, the last two compute-matched to the $L{=}48$ checkpoint. ``Mean'' is the unweighted mean of the eight language understanding benchmarks. \texttt{IFEval} (prompt-level strict accuracy) is reported separately and is excluded from the mean.}
\label{tab:general_capability}
\resizebox{\textwidth}{!}{%
\begin{tabular}{@{}l cccccccc c c@{}}
\toprule
\textbf{Model} & MMLU & ARC-E & ARC-C & HellaSwag & WinoG. & PIQA & BoolQ & LAMBADA & \textbf{Mean} & IFEval \\
\midrule
Base \texttt{Qwen3-1.7B} & 55.5 & 72.6 & 42.7 & 60.5 & 61.7 & 72.5 & 77.6 & 50.3 & 61.7 & 33.1 \\
\midrule
Curriculum, $L{=}8$ & 56.9 & 76.2 & 48.0 & 62.4 & 63.2 & 72.8 & 76.4 & 57.4 & 64.2 & 35.1 \\
Curriculum, $L{=}16$ & 57.1 & 75.7 & 45.9 & 62.5 & 62.4 & 73.3 & 77.1 & 58.0 & 64.0 & 39.0 \\
Curriculum, $L{=}32$ & 57.3 & 74.3 & 45.8 & 62.5 & 61.6 & 73.0 & 77.8 & 59.2 & 63.9 & 34.6 \\
Curriculum, $L{=}48$ & 56.1 & 74.4 & 45.6 & 64.0 & 59.8 & 73.3 & 77.2 & 56.5 & 63.4 & 41.2 \\
\midrule
No curriculum & 57.5 & 75.5 & 45.3 & 62.7 & 62.9 & 73.7 & 73.2 & 55.1 & 63.2 & 40.1 \\
Instruction-only & 56.4 & 75.7 & 48.3 & 64.0 & 60.4 & 73.3 & 79.2 & 56.4 & 64.2 & 46.4 \\
\bottomrule
\end{tabular}}
\end{table}

\section{Benchmark Descriptions}
\label{app:benchmark_descriptions}

Table~\ref{tab:benchmarks} summarizes the fourteen downstream reasoning benchmarks and the nine general-capability benchmarks: the reasoning type each targets, the answer format we score, the split or subset we evaluate, and the number of evaluation items $n$. The prompting protocol for each benchmark is given in Table~\ref{tab:eval_protocol}.

\begin{table}[p]
\centering
\footnotesize
\setlength{\tabcolsep}{4pt}
\renewcommand{\arraystretch}{1.5}
\linespread{0.9}\selectfont
\begin{tabular}{@{}>{\raggedright\arraybackslash}p{0.22\linewidth}>{\raggedright\arraybackslash}p{0.20\linewidth}>{\raggedright\arraybackslash}p{0.145\linewidth}>{\raggedright\arraybackslash}p{0.29\linewidth}r@{}}
\toprule
Benchmark & Reasoning type & Answer format & Split / subset & $n$ \\
\midrule
\multicolumn{5}{l}{\textbf{\emph{Reasoning benchmarks}}} \\
ProofWriter OWA$^\dagger$ \newline \citep{tafjord2021proofwriter} & Synthetic deductive & True/False/ Unknown & OWA, depths 0--5, label-balanced & 3000 \\
ProofWriter CWA \newline \citep{tafjord2021proofwriter} & Synthetic deductive & True/False & CWA, depths 0--5 & 3000 \\
ParaRules$^\dagger$ \newline \citep{clark2020ruletaker} & Synthetic deductive, paraphrased & True/False/ Unknown & ProofWriter OWA NatLang test, depths 0--5, label-balanced & 4566 \\
ZebraLogic-MC \newline \citep{lin2025zebralogic} & Constraint satisfaction & Multiple choice & \texttt{mc\_mode} & 1000 \\
PrOntoQA$^\dagger$ \newline \citep{saparov2023testing} & Synthetic deductive & True/False & generated, 8 depths up to 64 & 4000 \\
FOLIO \newline \citep{han2024folio} & First-order logic & True/False/ Unknown & validation & 203 \\
JustLogic \newline \citep{chen2025justlogiccomprehensivebenchmarkevaluating} & Knowledge-independent deductive & True/False/ Uncertain & depths 1--7 & 4897 \\
ProverQA \newline \citep{ICLR2025_862819c2} & First-order logic with distractors & 3-way multiple choice & dev, easy/medium/hard & 1497 \\
CLUTRR \newline \citep{sinha2019clutrr} & Inductive kinship & Relation label set & full & 1000 \\
LogicBench-BQA \newline \citep{parmar2024logicbench} & Single-rule deductive & Yes/No & BQA & 1512 \\
RuleTaker \newline \citep{clark2020ruletaker} & Synthetic deductive & True/False & test, depths D0--D5 & 1000 \\
Web of Lies$^\dagger$ \newline \citep{suzgun-etal-2023-challenging} & Truth-teller chains & Yes/No & generated, 8 depths up to 64, label-balanced & 4000 \\
LogiQA \newline (\citealp{liu2020logiqa};\newline \citealp{zhong2023agieval}) & Mixed (exam) & 4-way multiple choice & AGIEval split & 648 \\
NLGraph \newline \citep{wang2023nlgraph} & Graph reasoning & Open-ended (8 sub-tasks) & full & 1000 \\
\midrule
\multicolumn{5}{l}{\textbf{\emph{General-capability benchmarks}}} \\
MMLU \newline \citep{hendrycks2021mmlu} & World knowledge & 4-way multiple choice & test & 14042 \\
ARC-Easy \newline \citep{clark2018arc} & Science QA & 4-way multiple choice & test & 2376 \\
ARC-Challenge \newline \citep{clark2018arc} & Science QA & 4-way multiple choice & test & 1172 \\
HellaSwag \newline \citep{zellers2019hellaswag} & Commonsense inference & 4-way completion & validation & 10042 \\
WinoGrande \newline \citep{sakaguchi2020winogrande} & Commonsense inference & 2-way completion & validation & 1267 \\
PIQA \newline \citep{bisk2020piqa} & Physical commonsense & 2-way multiple choice & validation & 1838 \\
BoolQ \newline \citep{clark2019boolq} & Reading comprehension & Yes/No & validation & 3270 \\
LAMBADA \newline \citep{paperno2016lambada} & Reading comprehension & Last-word prediction & test (OpenAI) & 5153 \\
IFEval \newline \citep{zhou2023instruction} & Instruction following & Generation, prompt-level strict & full & 541 \\
\bottomrule
\end{tabular}
\vspace{-0.7em}
\caption{The fourteen downstream reasoning benchmarks and the nine general-capability benchmarks, with the number of evaluation items $n$. General-capability benchmarks use the default \texttt{lm-evaluation-harness} splits. \texttt{ProofWriter} appears twice because its open- and closed-world splits differ in label set. \texttt{NLGraph} is scored with the authors' per-task scorers, and we report the mean over its eight sub-tasks (Appendix~\ref{app:eval_methodology}). $^\dagger$Modified by us, as described in this appendix.}
\label{tab:benchmarks}
\end{table}

\paragraph{PrOntoQA.} The released \texttt{PrOntoQA} data stops at 4 to 5 hops, and the shipped generator cannot go further, because it samples a random ontology over 17 concept names and takes a random proof walk through it. We replace the ontology sampler with a chain of exactly $k+1$ concepts, script the walk to the unique $k$-step proof, and enlarge the vocabulary to 624 names and 274 adjectives. Rendering, inflection, parsing, distractors and label negation use the official code. An independent solver verifies every item for its label, for a proof depth equal to $k$, for consistency, and for distractor sign. The set covers eight depths, $k \in \{1, 2, 4, 8, 16, 32, 48, 64\}$, 500 items each, balanced on both the label and the query polarity.

\paragraph{ProofWriter OWA and ParaRules.} We use the published data but draw equal numbers of True, False and Unknown items within each official \texttt{QDep} stratum. Without this, the Unknown share falls from 69\% at depth 0 to 3\% at depth 5, because an unprovable question inherits the depth of its deepest failing search, so a model that always answers Unknown would score 66\% at depth 0 and 4\% at depth 5. Balancing fixes chance at 33.3\% at every depth. The closed-world split is already near 50/50 at every depth and needs no resampling. We apply the same balancing to \texttt{ParaRules}, whose items come from the \texttt{ProofWriter} open-world NatLang test split.

\paragraph{Web of Lies.} Our \texttt{Web of Lies} set is generated from the BBH \texttt{web\_of\_lies} template and its 50-name pool, as truth-value chains of $k \in \{1, 2, 4, 8, 16, 32, 48, 64\}$ speakers with 500 items per depth and exactly balanced Yes/No labels. Our solver reproduces all 250 published BBH items, all of which have $k{=}4$.

All other benchmarks are the published data as released. Source repositories: \texttt{ProofWriter} from AllenAI \texttt{proofwriter-dataset-V2020.12.3}; \texttt{FOLIO} from \texttt{yale-nlp/FOLIO}; \texttt{ZebraLogic} from \texttt{WildEval/ZebraLogic}; \texttt{RuleTaker} from \texttt{tasksource/ruletaker} (fallback \texttt{hitachi-nlp/ruletaker}); \texttt{CLUTRR} from \texttt{kendrivp/CLUTRR\_v1\_extracted} (fallback \texttt{kliang5/CLUTRR\_huggingface\_dataset}); \texttt{LogiQA} from \citet{zhong2023agieval}; \texttt{LogicBench} and \texttt{NLGraph} from their authors' releases.

\FloatBarrier

\section{Training Data Examples}
\label{app:examples}

We show generated training examples at three lookahead values to illustrate how problem complexity scales across curriculum stages. Each example consists of premises (rendered as conditional predicates), a query naming the start fact and goal, and a proof prefix ending where the model must predict. The model's target output is a single attribute (tokenized as one or more BPE tokens). Within a stage, lookaheads are sampled uniformly from 1 to the cap, so most stage-$L$ instances have lookahead below $L$; the examples below are chosen with lookahead equal to the cap.

\needspace{6\baselineskip}
\begin{examplebox}[unbreakable]{Stage cap $L{=}2$, example lookahead $\ell = 2$ (5 attributes)}
Suppose we have the following facts: If someone is itaema, then they are isum. If a person is ugphi, they are itaema. If Krystal is nwazao, then they are asico. If Krystal is ugphi, then they are nwazao. Given that Krystal is ugphi, and we want to prove Krystal is asico.

Proof: Krystal is ugphi. Krystal is \underline{\hspace{5em}}
\tcblower
\textbf{Answer:} \texttt{nwazao}
\end{examplebox}

\needspace{6\baselineskip}
\begin{examplebox}[unbreakable]{Stage cap $L{=}8$, example lookahead $\ell = 8$ (17 attributes)}
Suppose we have the following facts: If someone is kfusau, then they are kacown. If a person is aqyqoj, they are adawr. Everyone that is fucboo is nuuht. Everyone that is kacown is upocoo. If a person is yuekl, they are xwohoo. If Steven is buey, then Steven is zacipo. Everyone that is xuilat is wxilvu. Everyone that is ahxbwu is ezhol. If Steven is adawr, then Steven is fucboo. If someone is aqyqoj, then they are ornce. If a person is ecqni, they are ahxbwu. If someone is upocoo, then they are yuekl. If Steven is ezhol, then Steven is xuilat. Everyone that is nuuht is kfusau. If someone is ornce, then they are buey. If Steven is zacipo, then they are ecqni. Given that Steven is aqyqoj, and we want to prove Steven is xwohoo.

Proof: Steven is aqyqoj. Steven is \underline{\hspace{5em}}
\tcblower
\textbf{Answer:} \texttt{adawr}
\end{examplebox}

\needspace{6\baselineskip}
\begin{examplebox}{Stage cap $L{=}32$, example lookahead $\ell = 32$ (65 attributes)}
Suppose we have the following facts: If someone is pyaqau, then they are tsawob. If Jennifer is etuupu, then they are poqgje. If a person is vaorea, they are zfaqa. If Jennifer is dduosi, then they are biisi. Everyone that is saiaqt is heeru. If someone is ussdoi, then they are gidxio. Everyone that is fosin is miaize. If someone is xdilxo, then they are ivgau. Everyone that is azlwoo is inaba. Everyone that is hzueda is azlwoo. If Jennifer is owaosy, then they are ahuvyi. If Jennifer is inaba, then Jennifer is agez. If Jennifer is iklau, then they are pyaqau. If Jennifer is ahuvyi, then Jennifer is afueji. If a person is miaize, they are efeh. If Jennifer is wimeu, then they are taew. If Jennifer is efeh, then Jennifer is saiaqt. Everyone that is ssoene is ruosa. Everyone that is idudz is ezejoi. If someone is hijnua, then they are ughtuf. If Jennifer is biisi, then they are faxids. Everyone that is zguul is dpielu. If someone is elorbi, then they are ryuwqo. If Jennifer is aqico, then Jennifer is oresu. If Jennifer is ruosa, then they are hijnua. If Jennifer is afueji, then they are oxuogj. If Jennifer is itcox, then Jennifer is etuupu. If someone is gidxio, then they are raeesb. If a person is heeru, they are owaosy. If Jennifer is liiz, then they are elorbi. If Jennifer is uyqbop, then Jennifer is ekice. If a person is raeesb, they are uyqbop. If a person is umuf, they are iyeriu. If Jennifer is emaca, then they are ssoene. Everyone that is yoxo is fosin. Everyone that is ekice is zaxbom. If Jennifer is dpielu, then they are obaqka. If someone is apenii, then they are luanx. If someone is faxids, then they are ussdoi. Everyone that is sbool is wimeu. If Jennifer is xuuur, then they are emaca. If someone is oresu, then they are brapuz. If someone is obaqka, then they are yoxo. If someone is jiiel, then they are idudz. If someone is taew, then they are ufnsih. If a person is apenii, they are yutou. If Jennifer is ufnsih, then Jennifer is aqico. Everyone that is zaxbom is zguul. If someone is ivgau, then they are liiz. If Jennifer is noza, then Jennifer is azbigo. If Jennifer is brapuz, then they are jiiel. If a person is ezejoi, they are hzueda. If Jennifer is ikeyo, then Jennifer is sbool. Everyone that is agez is iklau. If Jennifer is yutou, then they are ikeyo. Everyone that is tsawob is umuf. If a person is luanx, they are xdilxo. If Jennifer is zfaqa, then they are neuklo. If Jennifer is oxuogj, then they are itcox. Everyone that is ughtuf is noza. If someone is ryuwqo, then they are enoxde. Everyone that is enoxde is dduosi. Everyone that is iyeriu is vaorea. If Jennifer is neuklo, then Jennifer is xuuur. Given that Jennifer is apenii, and we want to prove Jennifer is azbigo.

Proof: Jennifer is apenii. Jennifer is \underline{\hspace{5em}}
\tcblower
\textbf{Answer:} \texttt{yutou}
\end{examplebox}

\begin{figure}[H]
  \centering
  \includegraphics[width=\linewidth]{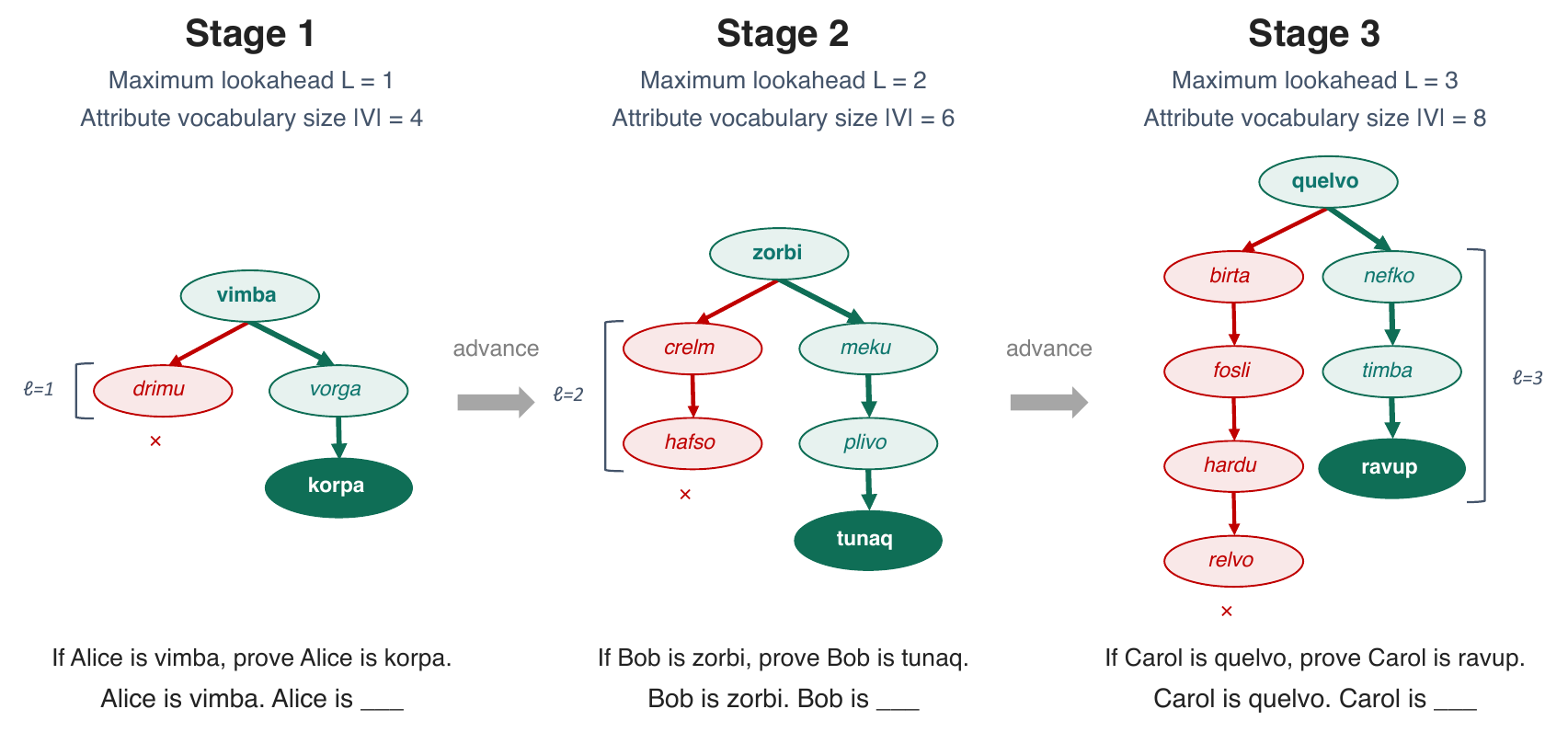}
  \caption{Curriculum over lookahead and attribute vocabulary size. The three stages use maximum lookahead $L{=}1, 2, 3$ and attribute vocabulary size $|\mathcal{V}_k| = 4, 6, 8$, respectively. The arrows indicate stage advancement. Each DAG is converted into natural-language premises; for example, the edge \texttt{vimba} $\to$ \texttt{vorga} reads ``If Alice is vimba, then Alice is vorga.'' The abbreviated examples below the graphs show the goal and proof prefix; the model predicts the next attribute at the blank. Green marks goal-reaching paths and red marks distractor paths. Neither the colors nor the graph diagrams are part of the model input.}
  \label{fig:curriculum_diagram}
\end{figure}

\end{document}